%% file: main_v5.tex
\documentclass[10pt,twocolumn]{article}

\usepackage[margin=0.72in]{geometry}
\usepackage{mathptmx}
\usepackage[T1]{fontenc}
\usepackage[utf8]{inputenc}
\usepackage{microtype}
\usepackage{amsmath,amssymb}
\usepackage{booktabs}
\usepackage{graphicx}
\usepackage{subcaption}
\usepackage{caption}
\usepackage{xcolor}
\usepackage{hyperref}
\usepackage[numbers,sort&compress]{natbib}

\hypersetup{colorlinks=true,linkcolor=blue,citecolor=blue,urlcolor=blue,
  pdftitle={Staged Semantic Distillation: Preserving Free-Form Semantics in a Low-Compute Non-LLM Surrogate of a Staged LLM Agent},
  pdfauthor={Zhiyang Niu, Huayuan Shen, Chuanjie Li, Mengmeng Zhao, Jiajia Zhang}}
\title{\vspace{-1.2em}\textbf{Staged Semantic Distillation: Preserving Free-Form Semantics}\\[3pt]
       \textbf{in a Low-Compute Non-LLM Surrogate of a Staged LLM Agent}}
\author{
Zhiyang Niu\thanks{Harbin Institute of Technology, Shenzhen, China. \texttt{2024311601@stu.hit.edu.cn}},
Huayuan Shen\thanks{Harbin Institute of Technology, Shenzhen, China. \texttt{25s151212@stu.hit.edu.cn}},
Chuanjie Li\thanks{Nanyang Technological University, Singapore. \texttt{LICH0073@e.ntu.edu.sg}},
Mengmeng Zhao\thanks{Pingan Group, China. \texttt{zmmuestc@163.com}},
Jiajia Zhang\thanks{Harbin Institute of Technology, Shenzhen, China. \texttt{zhangjiajia@hit.edu.cn}. Corresponding author.}
}
\date{}

\begin{document}
\maketitle
\vspace{-1.6em}

\input{sections_v5/00_abstract}
\input{sections_v5/01_intro}
\input{sections_v5/02_method}
\input{sections_v5/03_theory}

\input{sections_v5/04_interface}
\input{sections_v5/05_deployment}
\input{sections_v5/06_related}
\input{sections_v5/07_discussion}

\appendix
\small
\input{sections_v5/08_appendix}

\clearpage
\bibliographystyle{plainnat}
\bibliography{refs}

\end{document}

%% file: sections_v5/00_abstract.tex
\begin{abstract}
Staged LLM agents make market and social simulators realistic, but a single 102-day run of a 200-agent, ten-index
market simulator issues 203{,}100 language-model calls and returns one stochastic realization. Conventional
low-compute distillation sheds the property that makes these agents useful---their free-form intermediate state. We
present \emph{staged semantic distillation}, a low-compute method that preserves non-structured semantics in a non-LLM
student trained from one recorded trace. Each free-form teacher output is encoded once by a frozen encoder into a
1024-d vector; the student receives typed, stage-tagged vectors with stage-specific visibility, is supervised jointly on
all stages, and is trained under the deployment protocol with scheduled self-history. One TwinMarket trace yields a
13.2M-parameter deterministic student that trains in 1{,}658 seconds on one 8-GB consumer GPU and runs with zero teacher
or decoder calls and no text generation (0.14 ms per decision); its analysis stays close to the teacher's (cosine 0.943)
and swapping semantic inputs between agents moves 64.7\% of decisions against 12.8\% for a matched donor, so the
free-form channel is preserved and live.

We use the low-compute surrogate to study what the transfer carries. On the full test window it reaches 0.8347
balanced accuracy on the teacher-forced replay and 0.7783 [0.766,0.791] on the within-day deployment decoder; the
deployment result is that absolute balanced accuracy, and the five-seed recipe gain over a no-semantic retrain is
$+2.2$ points [1.21,3.14]. Separately, on round-one decisions, against the teacher's own 20-draw re-sampling level the
student's agreement gives $R=97.1\%$ [95.8,98.4] with the recorded analysis (balanced accuracy 0.9098) and 92.3\%
[89.0,95.4] with its own (0.7252); $R$ is a relative diagnostic, not an accuracy: a constant predictor reaches
53.4\%/63.6\% and $R$ can exceed 1. No later-round teacher replay exists, so the deployment decoder has no $R$. A
five-seed, 2$\times$3 content-by-schedule study isolates the deployment gain: scheduled self-history carries it
(+2.2 points [1.21,3.14] over no-semantic retrains; shuffled- and constant-content controls reproduce it), while
semantic content matters at the teacher-forced point. The practical design rule---invest in the deployment schedule, let content earn its
place at the teacher-forced operating point---is the study's main finding. Determinism makes the surrogate a cheap
instrument: a 20-seed, 200{,}000-decision sweep runs in 12 minutes and zero LLM calls, against about three hours and
$4.0\times10^5$ calls for the sampling teacher. The paper is a method-and-case-study contribution on one simulator and
one trace.
\end{abstract}

%% file: sections_v5/01_intro.tex
\section{Introduction}
\label{sec:intro}

LLM-agent simulators---simulators whose state is written and updated in free-form text by language-model agents---are
becoming a practical substrate for decision-making, multi-agent simulation, and policy evaluation
\citep{park2023generative,yang2025twinmarket,zhang2024finagent}. A market simulator of this kind lets a population of LLM
agents reason in text and trade from recorded news and prices. The approach is powerful but expensive: a single 102-day
run of a ten-index simulator with twenty agents per index (200 agents) issues 203{,}100 language-model calls and 691.5M
prompt tokens, and because the teacher samples its text and its actions, the run returns one realization of a
stochastic process.

The natural way to cut this cost is to distil the agents into a small non-LLM student. Existing low-compute
distillation, however, gives up exactly the property that makes these agents useful: their free-form intermediate
state. Standard distillation and imitation assume a single label, an action distribution, or a well-defined trajectory
\citep{hinton2015distilling,bengio2015scheduled,ross2011dagger,ho2016gail}; a staged LLM agent has none of these. Its
intermediate state is free-form text (the day analysis and the round rationale carry the decision-relevant knowledge),
its stages have different visibility, and deployment feeds the student its own previous outputs. Agent-distillation
keeps a small language model in the student \citep{kang2025agentdistill}, and tabular or MLP surrogates consume only
structured features; neither preserves free-form semantics in a non-LLM model. To our knowledge, staged semantic
distillation is the first low-compute method that keeps the free-form channel and runs the surrogate without any
language model at inference.

\paragraph{The method.}
\emph{Staged semantic distillation} has four components. (i) \emph{Semantic encoding}: each free-form teacher output is
encoded once by a frozen text encoder into a 1024-d unit vector; the student never parses or generates text.
(ii) \emph{A typed, stage-tagged interface}: stage vectors are typed tokens with stage-specific visibility, preserving
what each stage of the teacher could see. (iii) \emph{Joint stage supervision}: six stage heads are trained together, so
the analysis vector is both close to the teacher's and useful to later stages. (iv) \emph{Scheduled self-history}: with
probability $p$ the history slots contain the student's own generated vectors, training it on the deployment protocol.
On TwinMarket, a 200-agent, ten-index market simulator, one recorded 102-day trace yields a 13.2M-parameter
deterministic student that trains in 1{,}658 seconds on one 8-GB consumer GPU and runs the decision loop with zero
teacher or decoder calls, zero text generation, and 0.14 ms per decision; the frozen encoder is applied once, offline.
The method is the main contribution. We offer it as a recipe for staged LLM agents whose intermediate outputs are text,
and validate it on one instantiation; Section~\ref{sec:discussion} states the conditions for transferring it.

\paragraph{The study it enables.}
A deterministic surrogate at this cost is also a measurement instrument, so we can ask what of the staged process
transfers and which component carries it. We anchor the study on the teacher itself: the trace is one sample of a
stochastic teacher, so we replay the teacher 20 times on each round-one test context and measure the student's agreement
relative to the teacher's own re-sampling level, a replay-calibrated diagnostic ratio $R$ (not an equivalence claim).
The surrogate's determinism makes component and seed studies cheap enough to isolate the mechanism. The study finds
that the deployment gain is carried by scheduled self-history: content-shuffled and content-constant controls reproduce
the within-day deployment accuracy, and removing the schedule costs 5.9 points with matched content and 2.3--3.2 points
with the controls. Semantic content matters where it can be read, at the teacher-forced point, where every arm receives
the teacher's recorded analysis. The resulting design rule for low-compute semantic-preserving surrogates---invest in the
deployment schedule, let content earn its place at the teacher-forced operating point---is the study's main practical
finding.

\paragraph{What we establish.}
\begin{itemize}
\item \textbf{A low-compute method that preserves non-structured semantics in a non-LLM student.} The four-component
recipe above, instantiated on TwinMarket: a 13.2M-parameter deterministic student, 1{,}658 s on one consumer GPU, zero
teacher or decoder calls, 0.14 ms per decision, and a free-form channel that stays close to the teacher's (cosine 0.943)
and is functionally live (donor swap 64.7\% vs.\ 12.8\%). This capability---a non-LLM surrogate that keeps the staged
free-form information---is what existing low-compute distillation does not offer.
\item \textbf{A replay-calibrated study of what the transfer carries.} 0.8347 balanced accuracy teacher-forced and
0.7783 on the deployment decoder; $R=97.1\%$ [95.8,98.4] and 92.3\% [89.0,95.4]. A five-seed, 2$\times$3
content-by-schedule design isolates the mechanism: scheduled self-history carries the deployment gain (+2.2 points
[1.21,3.14] over no-semantic retrains; shuffled/constant-content controls reproduce it), while content matters for
teacher-forced fidelity. This is a transferable design rule, not just a number.
\item \textbf{A cheap, reproducible instrument.} Determinism under a fixed simulator seed makes repeated evaluation
practical: a 20-seed, 200{,}000-decision sweep takes 12 minutes and zero LLM calls, against about three hours and
$4.0\times10^5$ calls for the sampling teacher. The surrogate does not replace the first trace; it removes the teacher
from every subsequent run.
\item \textbf{A specified recipe and implementation.} The architecture, stage interface, visibility matrix, losses,
schedule and training budget are documented, and the semantic channel is measured with read-outs, head refits and input
interventions, so the method can be applied when a new staged agent trace becomes available.
\end{itemize}

\paragraph{Scope.}
The empirical study is a case study on one simulator, one recorded trace, one five-month market regime and one student
family; it demonstrates the method and its transfer analysis on this instantiation and does not claim generality across
simulators. Section~\ref{sec:method} presents the method, Section~\ref{sec:theory} the replay-calibrated protocol,
Section~\ref{sec:interface} the transfer study, Section~\ref{sec:deployment} the surrogate and its cost/closed-loop
properties, Section~\ref{sec:related} related work, and Section~\ref{sec:discussion} the discussion.

%% file: sections_v5/02_method.tex
\section{Staged Semantic Distillation}
\label{sec:method}

\begin{figure*}[t]
\centering
\includegraphics[width=0.98\textwidth]{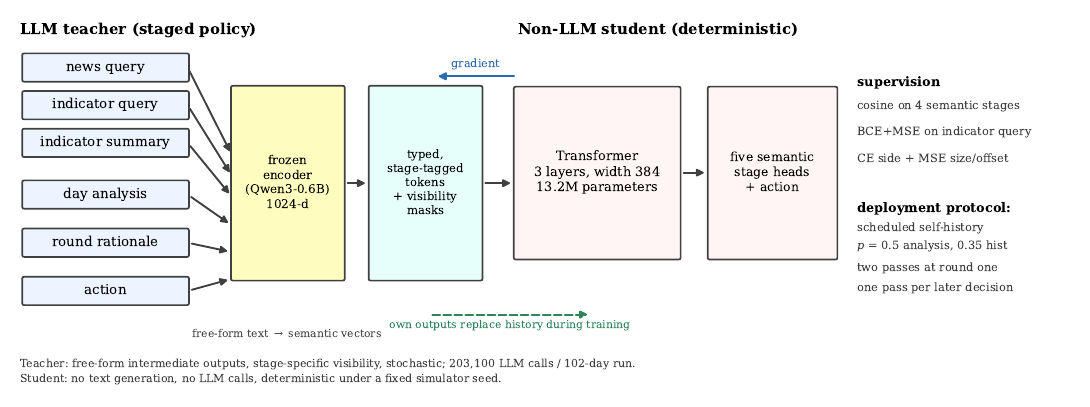}
\caption{Staged semantic distillation. Left: the teacher's staged policy writes free-form outputs at five semantic
stages and then acts; the stages have different visibility, and the recorded text is the intermediate state to
preserve. Middle: each output is encoded once by a frozen text encoder into a 1024-d vector and exposed to the student
as a typed, stage-tagged token with a visibility mask. Right: a 13.2M-parameter Transformer is trained with joint stage
supervision and scheduled self-history and runs deterministically with no text generation and no teacher or decoder
calls at inference.}
\label{fig:method}
\end{figure*}

\paragraph{Setting.}
An \emph{LLM-agent market simulator} is a generative process in which a population of language-model agents maintains a
shared environment state: each round an agent reads a context $x_t$ (persona, market news, indicator summaries, account
state, its own previous outputs), writes free-form text, and acts; actions update the environment. Our instantiation is
TwinMarket, a 200-agent, ten-index market simulator: the decision process has six stages (news query, indicator query,
indicator summary, day analysis, round rationale, buy/sell action), the trace contains the teacher's text at every
stage, and the simulated market path is the world state the agents produce. We distil the agents' staged policy, not the
environment: the order book and price formation stay with the simulator. The task is to learn a non-LLM deterministic
surrogate that reproduces this staged process, at the level of decision-relevant semantic content, from the trace alone
and without the teacher at inference. Figure~\ref{fig:method} gives the pipeline.

\paragraph{The record.}
The trace covers 102 trading days, twenty agents per index on ten Chinese industry indices (200 agents), and five rounds
per day (102{,}000 agent-decisions); producing it required 203{,}100 language-model calls and 691.5M prompt tokens. Days
are split 86/6/10 into train, validation and test; the round-one test set contains 1{,}998 labelled decisions and the
full test set is used for the autoregressive and closed-loop evaluations.

\paragraph{Semantic encoding, interface, and visibility.}
Each free-form output is passed once through a frozen text encoder into a 1024-d unit vector; the encoder is applied
offline to the trace, so the decision loop contains no language model. The student sees only fixed semantic vectors. It
never parses or generates this text: the interface is typed and stage-tagged, but its content is learned vectors. Every
stage has a type tag and an attention prefix; appendix~\ref{sec:mask} gives the exact visibility matrix. In words, the
news-query head sees the persona and its query token; the indicator-query head sees persona, previous-day history and
its query token; the indicator-summary head adds the generated indicator query and indicator values; the analysis head
sees persona, the two retrieved-news tokens, the indicator summary, account state, extra context and previous-day
history, but never the current day's rounds; the rationale and action heads see the full context, and the action head
additionally reads its hidden state concatenated with the generated analysis and rationale vectors. Retrieved news is a
precomputed input from a fixed corpus; the news-query head is supervised but does not drive retrieval in this
implementation. At deployment the deployed student generates all six stages itself.

\paragraph{Joint stage supervision.}
The deployed student is a Transformer encoder with three layers, width 384, eight attention heads and 13{,}175{,}439
parameters, with its own output head per stage. Training minimises cosine losses between predicted and recorded stage
vectors on the four semantic stages, $0.5\cdot$BCE $+\,0.05\cdot$MSE on the indicator query, cross-entropy on the action,
and $0.01\cdot$MSE on order size and offset; all stages are supervised jointly, so the analysis head produces a vector
that is both close to the teacher's and useful to later stages. Each stage head writes back into the same 1024-d space
the inputs use.

\paragraph{Deployment protocol and scheduled self-history.}
At deployment the teacher's analysis and rationale are unavailable: the student must generate its own semantic history
and condition on it in later rounds. We therefore train with scheduled self-history: with probability $p_{\mathrm{self}}$
a history slot contains the student's own generated vector and with probability $1-p_{\mathrm{self}}$ the recorded teacher
vector. The deployed checkpoint uses $p_{\mathrm{self}}=0.5$ for the analysis slot and $0.35$ for the rationale and
action-history slots. The schedule mixes recorded and self-generated history, so training contains both sources; at
inference the history is self-generated. Round one has no previous round, so its history slot is empty at training and
test time; the current day-analysis vector is what the schedule replaces.

\paragraph{Inference.}
The action and rationale stages run in a single forward pass per decision; the day analysis needs one additional pass per
agent-day at round one, so a round-one decision takes two passes and a later-round decision one. The student writes no
text and calls no teacher or decoder language model; the frozen encoder is used only offline when the trace is
constructed. The same checkpoint serves the one-step replay, the within-day autoregressive rollout and the closed-loop
simulator.

\paragraph{Two diagnostic information sets.}
We evaluate two round-one operating points on the same 1{,}998 test decisions. \textbf{(i) Teacher analysis}: the
recorded analysis vector occupies the analysis slot (teacher-forced, as during training), the round-one rationale slot is
empty, and the student generates the rationale; this is the most teacher-like point and matches a replay that keeps the
recorded analysis fixed. \textbf{(ii) Self-generated analysis}: the slot is first emptied, the student generates its own
day-analysis vector, and the action is read after re-injecting it; this matches a replay that generates a fresh analysis
before deciding and is the deployment-relevant round-one point. The self-analysis comparison uses the 1{,}982 rows in
complete five-round agent-days and recomputes the replay reference on those rows (0.7858). A third measurement, the
\emph{no-semantic-history retrain}, uses the identical architecture, data, split and optimiser but zeros the
semantic-history tensor and mask during training and evaluation; because those slots are absent it also disables
scheduled self-history, so the contrast bundles the slots, their masks and the schedule, and the no-schedule cells in
Section~\ref{sec:interface} separate the schedule component. We refer to the teacher-forced one-step replay as P1 and the
within-day autoregressive rollout as P2.

%% file: sections_v5/03_theory.tex
\section{Evaluating a Deterministic Student of a Sampling Teacher}
\label{sec:theory}

\paragraph{The recorded action is one sample.}
The teacher is stochastic: at a given context it samples its text and its action. The trace contains one draw, so
treating the recorded action as noiseless ground truth charges the student for the teacher's sampling noise. We
therefore replay the teacher on the recorded information and ask how often a fresh draw reproduces the recorded
decision. This is the teacher's \emph{own re-sampling level}, the natural reference for a deterministic surrogate whose
output is used as one sample of the teacher's decision distribution.

\paragraph{The re-sampling level and sample fidelity.}
Let $p_T(\cdot\mid x)$ be the teacher's conditional action distribution at context $x$ and $\pi_a$ the class
frequencies. For a fresh teacher draw at the recorded context, the expected agreement with the recorded action in
balanced-accuracy space is $S_{\mathrm{bal}}=\tfrac12\,\mathbb{E}_x[\sum_a p_T(a\mid x)^2/\pi_a]$. For a deterministic
student $f$ whose output is a point mass at one class, the analogous quantity is
$A_{\mathrm{bal}}(f)=\tfrac12\,\mathbb{E}_x[p_T(f(x)\mid x)/\pi_{f(x)}]$, which equals the student's standard balanced
accuracy against the recorded labels, $\tfrac12(\mathrm{TPR}+\mathrm{TNR})$, the estimator reported in the tables.
\emph{Sample fidelity} is the ratio $R(f)=A_{\mathrm{bal}}(f)/S_{\mathrm{bal}}$: the student's balanced accuracy divided
by the teacher's re-sampling level in the same space. Both quantities are estimated on the same recorded decisions, so
the recorded action's sampling noise enters neither side differentially. The denominator is recomputed on the rows of
each comparison: 0.9368 on the 1{,}998 teacher-analysis rows and 0.7858 on the 1{,}982 self-analysis rows (0.7860 on the
full 1{,}998). We report the measured ratio with a cluster-bootstrap interval.

\paragraph{Two information sets, matched to the two student operating points.}
The comparison is meaningful only when the student and the re-sampled teacher see the same information. In condition
\textbf{A} the recorded analysis is kept fixed and only the decision is re-sampled; this matches the teacher-forced
operating point, with the recorded analysis in the current slot and the round-one rationale history empty. In condition
\textbf{B} a fresh analysis is generated before each decision; this matches the deployment-relevant self-analysis
operating point, in which the student first generates its own day-analysis vector and then acts.

\paragraph{Estimating the reference.}
The trace was generated on 2026-11-06; both replay conditions were run on 2026-11-10 with the same API model,
\texttt{deepseek-chat}, which the provider does not version-pin and for which no immutable snapshot identifier is
exposed. A version-sensitivity check re-replays 128 of the contexts on 2026-11-14, four days after the replay, with the same API: the re-sampling
level moves by $-0.36$ points in condition A (95\% interval $[-1.24,+0.44]$) and $+0.26$ points in B
($[-1.73,+2.26]$), i.e., within replay noise, so the denominator is stable over this interval even though the provider
does not pin a snapshot. For each of the 1{,}998 round-one test contexts we replay the teacher 20 times (a few contexts have fewer
successful draws; mean 19.97): 39{,}960 decision calls in condition A and 79{,}920 calls in condition B, where each draw
additionally generates the fresh analysis. The resulting conditionals estimate the two re-sampling levels of
Table~\ref{tab:ceiling}. Conditions A and B differ by construction (0.9368 vs.\ 0.7860): A supplies the recorded
analysis and B requires a fresh one, so they are different information-specific quantities, not two estimates of one
number. Within each condition the balanced- and accuracy-space estimators agree within 0.0003. Ratios, accuracy
differences and distributional metrics use a two-level bootstrap: date$\times$asset clusters (100 clusters, 1{,}000
replicates) are resampled, and the per-context replay draws are redrawn from $\hat p$ within each replicate, so the
intervals include context and replay-sampling noise. A date-cluster re-analysis over the ten dates gives $R$ intervals
[96.2,97.9] in A and [90.1,94.3] in B, no wider than the two-level intervals, so the ten-date structure does not change
the conclusions.

\begin{table}[t]
\centering
\footnotesize
\caption{Teacher re-sampling level on the round-one test decisions ($n=1{,}998$; 20 replays per context), in
balanced-accuracy and accuracy space; the balanced 95\% cluster-bootstrap interval is [0.9296,0.9448] for A and
[0.7729,0.7998] for B. Condition A is the denominator of the teacher-forced (P1) fidelity ratio; condition B the
denominator of the round-one self-analysis ratio.}
\label{tab:ceiling}
\setlength{\tabcolsep}{4pt}
\begin{tabular}{lcc}
\toprule
Information set & Parity bal. & Parity acc. \\
\midrule
A: recorded analysis & 0.9368 & 0.9370 \\
B: fresh analysis & 0.7860 & 0.7863 \\
\bottomrule
\end{tabular}
\end{table}

\paragraph{Scoring the predicted distribution.}
The side head also emits a probability $q(a\mid x)$, which directly approximates the teacher's conditional action
distribution. We score $q$ against the replayed conditional $\hat p$ with the binary Brier score, total variation, KL
divergence, expected calibration error in ten equal-mass bins, the fraction of contexts where the student's and the
resampled teacher's most likely actions coincide, and the NLL of the recorded action under $q$; full definitions and the
uncertainty caveat are in the appendix.

\paragraph{What the protocol measures.}
$R$ is a diagnostic comparison: it says how the student's agreement with the recorded action compares with the
agreement a fresh teacher draw would achieve. It is not an equivalence claim. The reference is a same-API re-sample of
\texttt{deepseek-chat}, which the provider does not version-pin; a constant predictor reaches $R=53.4\%$ in A and
$63.6\%$ in B, and $R$ can exceed 1, so we report excess over chance and absolute balanced accuracy alongside it. The
protocol is complemented by the distributional scores above and a strong tabular deterministic baseline
(gradient-boosted trees on recorded features); it is behavioural and information-set specific, and the closed loop is
evaluated separately on market properties (Section~\ref{sec:deployment}).

%% file: sections_v5/04_interface.tex
\section{What the Method Transfers: A Replay-Calibrated Study}
\label{sec:interface}

This section uses the surrogate as an instrument to study what the staged transfer carries and where. All comparisons
use the same labels and split; row sets and readers are stated per comparison, and the appendix accounts for the
differences. We report the teacher-forced operating point, the deployment point, and then the component study that
isolates what carries the deployment gain.

\subsection{Teacher-forced fidelity and the semantic channel}
With the teacher's recorded analysis in the input, the student generates its own stage vectors and action. The generated
analysis is close to the teacher's (mean cosine 0.943, sd 0.031, 5--95\% [0.885,0.976] on the 1{,}998 round-one rows);
the indicator summary reaches 0.93, the rationale 0.84, the auxiliary news query 0.80, and we do not read a stage
ordering into cosines measured on different stage outputs. This point supplies an oracle input: it measures whether the
typed vectors can carry the decision, not whether the student can produce that input at deployment. The teacher's own
re-sampling level is 0.9368 balanced accuracy; the deployed student reproduces the recorded decisions at 0.9098, a sample
fidelity of $R=97.1\%$ [95.8,98.4], a 2.7-point gap. The interface checkpoint's seed-0 run reaches 0.9242
($R=98.7\%$ [97.4,99.8]), but that round-one advantage does not carry to the full window (0.8192 one-step full-test
balanced accuracy against the deployed model's 0.8347), so the round-one ratio is not a predictor of deployment accuracy.
In the self-generated analysis point the reference is 0.7858 on the matched 1{,}982 rows and the deployed checkpoint
reaches 0.7252 ($R=92.3\%$ [89.0,95.4]); five deployed-recipe seeds average 0.7339 and five no-semantic retrains 0.7522.
Table~\ref{tab:fidelity} collects the primary results: the deployment block (P2) and the round-one diagnostic block.

\begin{table}[t]
\centering
\footnotesize
\caption{Primary results. The top block is the deployment result (P2 within-day autoregressive decoder) and the
five-seed recipe contrast; it is what the deployment claims rest on. The bottom block is the round-one sample-fidelity
diagnostic: $R$ is the student's balanced accuracy relative to the teacher's own re-sampling level, not an accuracy---a
constant predictor reaches 53.4\%/63.6\% on the same scale and $R$ can exceed 1. No later-round teacher replay exists,
so the deployment decoder has no $R$; the schedule/content decomposition is in Table~\ref{tab:contentsched}.}
\label{tab:fidelity}
\scriptsize
\setlength{\tabcolsep}{1.5pt}
\begin{tabular}{lccc}
\toprule
Quantity & Value & Diagnostic & Note \\
\midrule
\multicolumn{4}{l}{\emph{Deployment (P2, full test window)}}\\
Deployed (5-seed mean) & \textbf{0.7776} & -- & 0.7712--0.7825 \\
No-semantic retrain (5 seeds) & 0.7552 & -- & \\
Deployment gain & $+2.2$ & [1.21,3.14] & rounds 2--5 \\
Schedule removal (P2) & 5.9 / 2.3 / 3.2 & -- & Table~\ref{tab:contentsched} \\
\midrule
\multicolumn{4}{l}{\emph{Round-one diagnostic (not a deployment predictor)}}\\
Teacher-forced (P1) & 0.9098 & $R=97.1\%$ [95.8,98.4] & denom.\ 0.9368 \\
Self-analysis & 0.7252 & $R=92.3\%$ [89.0,95.4] & denom.\ 0.7858 \\
Interface ckpt (s0) & 0.9242 / 0.7526 & 98.7\% / 95.8\% & teacher / self \\
\bottomrule
\end{tabular}
\end{table}

The semantic channel is functionally live, not only cosine-close. An opposite-side donor swap moves 64.7\% of decisions
against 12.8\% for a same-side donor, clearing the analysis slot drops the round-one decision to 0.6149, and a read-out
of the generated analysis recovers most of the decision signal (0.7516 against the deployed head's 0.7252). The gap to
the teacher's recorded-analysis read-out (0.9076) is 15.6 points, and cross-reader cells place that distance in the
vectors rather than in the reader; a fresh head on the same vector recovers 2.1 points at round one. Substituting the
teacher's recorded indicator query and summary for the student's own leaves the round-one result at 0.7258 versus 0.7268,
so the upstream information-set mismatch is second-order and the remaining distance is at the analysis slot. The full
read-out, head-refit and upstream tables are in the appendix.

\subsection{Distributional fidelity}
\begin{figure*}[t]
\centering
\includegraphics[width=\textwidth]{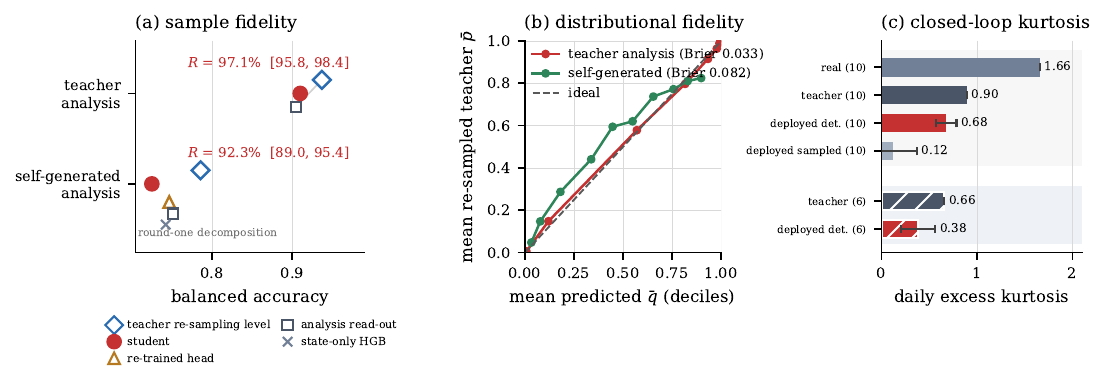}
\caption{(a) Teacher re-sampling level (open diamond) and deployed-student balanced accuracy (filled circle) for the two
information sets; the self-analysis row also shows decomposition references (fresh heads on the same generated vector,
an HGB read-out of the generated analysis, and a state-only HGB baseline), which consume rather than generate the
analysis vector. (b) Reliability of the deployed side probability against the re-sampled teacher probability
(Brier 0.033 with the teacher's analysis, 0.082 with its own). (c) Closed-loop daily excess kurtosis over the 102-day
window, mean $\pm$ sd across three simulator seeds; the index set is labelled and sampling collapses the tails.}
\label{fig:v5}
\end{figure*}

Scored against the replayed conditional, the deployed side probability reaches a binary Brier score of 0.033 and mode
agreement 0.931 at the teacher-analysis point, and 0.082 and 0.770 at the self-analysis point;
Figure~\ref{fig:v5}(b) shows the reliability curve. The probability is close to the replayed conditional in the
teacher-analysis condition and less sharp in the self-analysis condition.

\subsection{The component study: what carries the deployment gain}
Reproducing individual decisions is necessary but not sufficient, so we evaluate the recipe on the full test window and
in closed loop. The full-schedule model reaches 0.8347 balanced accuracy one-step and 0.7783 autoregressively. A
five-seed no-semantic retrain reaches 0.7848 one-step and 0.7552 autoregressively; the arm-mean deployment gain is
$+2.2$ points [1.21,3.14], rounds 2--5. The rest of this subsection isolates what produces that gain.

\paragraph{Typed interface versus a flat sequence model.}
A flat-token transformer with the same input tokens, no stage tags, visibility masks, stage heads or stage losses, the
same data, split, optimiser and budget (6 layers, 13.35M active parameters) reaches 0.8177 one-step balanced accuracy
against the staged model's 0.8347, a paired difference of $+1.70$ points [0.94,2.54] and 12.0\% decision disagreement.
The flat model cannot run the self-analysis protocol because it has no stage heads, so this is a one-step pipeline
comparison; it shows that the staged pipeline adds decision accuracy at the same inputs and budget.

\paragraph{Content versus schedule.}
A five-seed, 2$\times$3 design crosses content (agent-matched, within-date shuffled, constant) with the self-history
schedule (on, off). With the schedule on, the cells for agent-matched, shuffled and constant content read 0.7783,
0.7816 and 0.7857 on the rollout, so content changes it by at most 0.74 points in point estimate and the intervals
include zero. Removing the schedule costs 5.9 points with matched content (0.7783 to 0.7193), 2.3 points with shuffled
content (0.7816 to 0.7591) and 3.2 points with constant content (0.7857 to 0.7534), where the semantic schedule is
inert and the remaining schedule is the action-history schedule; every interval excludes zero. Agent-matched content matters at the
teacher-forced point: the shuffled/constant controls fall 4.0--4.5 points below the deployed arm in one-step accuracy
(seed-level details in Appendix Table~\ref{tab:controls}), but every arm receives the teacher's recorded analysis there and the
round-one history slots are empty, so this is a training-time association with contentful history, not a test-time
content effect. The design rule follows: the deployment decoder is carried by scheduled self-history, and the role of
agent-matched content is the teacher-forced operating point.
Table~\ref{tab:contentsched} gives the cells and the appendix reports the seed-level and interval details.

\begin{table}[t]
\centering
\footnotesize
\caption{Content-by-schedule design on the within-day deployment decoder (five deployed-recipe seeds, three seeds per
control cell). The constant row fixes only the semantic-history content; the current context, persona/news/state inputs
and, with the schedule on, the action-history schedule remain, so it is not an empty-input control. Removing the
schedule is the active ingredient in every content cell; with the schedule on, content changes the decoder by at most
0.74 points (intervals include zero).}
\label{tab:contentsched}
\setlength{\tabcolsep}{4pt}
\begin{tabular}{lccc}
\toprule
Content & Schedule on & Schedule off & Difference \\
\midrule
Agent-matched & 0.7783 & 0.7193 & 5.9 \\
Within-date shuffled & 0.7816 & 0.7591 & 2.3 \\
Constant & 0.7857 & 0.7534 & 3.2 \\
\bottomrule
\end{tabular}
\end{table}

\paragraph{Recipe-level and round-one results.}
The full schedule is the best of the recipe variants in balanced accuracy; the stage-loss and schedule extent
components are reported as a package, with a 2$\times$2 interaction whose difference-in-differences interval includes
zero. Round one is a distinct operating point: the retrained students are stronger there while the deployed gain appears
from round two on, and we report it separately in the appendix rather than mixing it with the deployment result. A
tied-head, no-action-bypass variant reaches the highest round-one fidelity (97.9\%/93.4\%) but over-shoots the
teacher's ten-index closed-loop tail level (2.42 vs.\ 0.90), which is why the full-schedule model is the one reported as
the system.

%% file: sections_v5/05_deployment.tex
\section{Deployment: A Deterministic Simulator with No Teacher Calls in the Loop}
\label{sec:deployment}

The deployed surrogate is a 13.2M-parameter model (52.7 MB) trained in 1{,}658 seconds on one 8-GB consumer GPU. Once
trained, it replaces every teacher and decoder call in the decision loop: the loop issues zero teacher or decoder
language-model calls, generates zero text, and performs one forward pass per decision after the day's analysis vector
has been generated. The frozen 0.6B encoder is not in the loop; it was applied once, offline, to construct the trace. At
batch 256 the policy runs at 0.14 ms per decision (0.59 ms at batch 20, 10.7 ms at batch 1), against 203{,}100 LLM calls,
691.5M prompt tokens and about 93 minutes for the recorded teacher run.

\paragraph{Accuracy.}
Over the full test window the student reaches 0.8347 balanced accuracy [0.826,0.843] one-step and 0.7783
[0.766,0.791] on the within-day autoregressive rollout. There is no later-round teacher replay, so the deployment
decoder has no $R$; the round-one $R$ values below are diagnostics relative to the teacher's re-sampling level (a
constant predictor reaches 53.4\%/63.6\% on that scale, and $R$ can exceed 1), not deployment accuracy. The
autoregressive decoder is the deployment decoder; the one-step value supplies the teacher's recorded analysis at every
round and is an oracle-input upper operating point. Five deployed-recipe seeds average 0.8311 (one-step) and 0.7776
(autoregressive), against 0.7848 and 0.7552 for five no-semantic retrains, so the arm-mean gains are $+4.63$ [3.6,5.6]
and $+2.24$ [1.21,3.14].

\paragraph{Closed-loop behaviour.}
A simulator is deployed to produce markets, so we ran the full 102-day loop with three simulator seeds. The student is
thinner than the teacher (2{,}322 vs.\ 13{,}484 lots per index per day) and the loop is initialised from the teacher's
account state on a window in which 86 of 102 days are training days, so the macro rows are descriptive rather than
size-matched. We report them as context: the deterministic surrogate's closed-loop daily excess kurtosis over ten
indices is 0.68 against 0.90 for the teacher path and 1.66 for the real market, and sampling size and offset from the
same checkpoint collapses the tails to 0.12; on the six indices whose teacher path is normal, the student's kurtosis is
0.38 against 0.66 and its volatility 163 against 168 bp, both inside the teacher path's date-block intervals. The
appendix gives the protocol and the held-out-day checks; the macro comparison is a descriptive consistency check, not a
fidelity fraction.

\paragraph{A twenty-seed sweep.}
Because the surrogate is deterministic under a fixed simulator seed, repeated evaluation is cheap. A 20-seed sweep over
the ten held-out days runs 200 realizations and 200{,}000 decisions in 12 minutes wall-clock on three CPU processes,
with zero LLM calls; seed-level mean excess kurtosis is $-0.29\pm0.28$ and volatility $146.9\pm7.5$ bp. This is a
10-day, one-index-per-run estimand, reported to show what simulator-side variation looks like at that scale; the
sampling teacher would need about three hours and $4.0\times10^5$ calls for the same twenty realizations. The full study
is in the appendix.

\paragraph{Reproducibility and cost.}
A deterministic agent policy with a fixed simulator seed re-runs exactly; repeated evaluation of the same rows is not
bit-reproducible under bf16 (0.8344 vs.\ 0.8347), so the third decimal is numerical noise. The system runs offline and
privately. The first run is not cheaper than the teacher: it requires the 203{,}100-call trace (about 93 min), one
encoder pass over the trace text, 1{,}658 s of training and about 2 min of simulation. The amortisation is on re-runs:
each additional 102-day deterministic realization costs about 2 min and zero LLM calls, against about 93 min and
203{,}100 calls for another teacher run; the re-sampling protocol is evaluation-only (119{,}880 replay calls).

%% file: sections_v5/06_related.tex
\section{Related Work}
\label{sec:related}

\paragraph{LLM agents and social simulation.}
LLM agents populate generative-agent societies \citep{park2023generative} and financial simulators
\citep{yang2025twinmarket,zhang2024finagent}. They produce reasoning and behaviour that rule-based agents do not, at the
cost of per-decision language-model inference and stochastic, non-reproducible runs. TwinMarket is a representative
staged market simulator of this line; our work distils its agents' policy, leaving the order book and price formation
with the simulator.

\paragraph{Distillation and imitation.}
Classical distillation transfers a teacher's output distribution to a student \citep{hinton2015distilling}; sequence-level
distillation and DAgger train on teacher states or on-policy rollouts \citep{bengio2015scheduled,ross2011dagger}, and
imitation learning and adversarial imitation learn policies from demonstrations \citep{ho2016gail}. These methods assume
that the teacher's output is a label, a distribution over actions, or a well-defined trajectory. A staged LLM agent
additionally exposes free-form intermediate text, stage-specific visibility, and a deployment protocol in which its own
outputs become history. The closest LLM-side lines make the contrast concrete. Agent-distillation distils an LLM agent
into small language models that still run a language model with retrieval and code tools at inference
\citep{kang2025agentdistill}; it targets action-level behaviour and keeps the student inside the LM family.
Continuous-rationale distillation compresses chain-of-thought into a continuous space by self-distillation
\citep{shen2025codi}, and Coconut trains LMs to reason in a continuous latent space \citep{hao2024coconut}; both keep the
student a language model and use the continuous space to shorten reasoning, not to replace the LM in the deployment loop.
World-model self-distillation trains a world model through self-distillation \citep{stapf2026worldmodel}, and
laptop-scale agent-society simulation reduces the cost of running LLM agents \citep{itkin2026poorman}; neither produces a
non-LLM student that carries a staged agent's free-form intermediate state. Our setting differs in the object being
distilled: a staged agent's policy, distilled from one trace into a non-LLM student whose inference loop contains no
language model while the free-form semantic channel is retained.

\paragraph{World models and agent surrogates.}
World-model distillation and surrogate simulators replace expensive simulators with learned models
\citep{stapf2026worldmodel,itkin2026poorman}; staged agent distillation and imitation from a single trace have recently
been studied for LLM agents \citep{kang2025agentdistill,shen2025codi}. Our method preserves the staged textual
intermediate state instead of compressing it away, and the deployment loop is decoder-free.

\paragraph{Evaluating a deterministic student of a sampling teacher.}
A deterministic student of a stochastic teacher is evaluated against a moving target. Our re-sampling reference is
related to noise ceilings, inter-rater agreement and test--retest reliability; we adapt these ideas to a staged LLM agent
by replaying the teacher on the recorded contexts in two information sets and reporting the student's agreement relative
to the teacher's own re-sampling level, with distributional and macro read-outs.

\paragraph{Market stylised facts.}
Heavy tails, volatility clustering and the absence of simple return autocorrelation are robust features of financial
returns \citep{cont2001empirical}, and market impact and liquidity have well-known structural models
\citep{kyle1985continuous,almgren2001optimal,glosten1985bid}. We use these statistics as a descriptive check of the
deployed surrogate: deterministic decoding preserves the heavy tails that sampling removes, so the stylised facts are
not an accident of the evaluation window.

\paragraph{Positioning.}
The paper contributes (i) a low-compute method that preserves free-form semantics in a non-LLM student, (ii) a
replay-calibrated study of what the transfer carries, (iii) a cheap deterministic instrument for repeated simulator
evaluation, and (iv) a specified implementation. The internal controls are not a substitute for an external distillation
baseline; the paper reports a method-and-case-study contribution on one simulator and one trace.

%% file: sections_v5/07_discussion.tex
\section{Discussion and Conclusion}
\label{sec:discussion}

\paragraph{What the paper proposes.}
We proposed staged semantic distillation, a low-compute method that turns a staged LLM agent into a non-LLM
deterministic surrogate while preserving its free-form intermediate semantics. The method encodes every free-form
teacher output once into a 1024-d semantic vector, exposes the staged process through a typed, stage-tagged interface
with stage-specific visibility, supervises all stages jointly, and trains the student under the deployment protocol with
scheduled self-history. On TwinMarket, a 200-agent, ten-index market simulator, one recorded trace yields a
13.2M-parameter student that trains in 1{,}658 seconds on one 8-GB consumer GPU and runs the decision loop with zero
teacher or decoder calls and zero text generation.

\paragraph{What the study shows.}
The surrogate reaches 0.8347 balanced accuracy at the teacher-forced operating point and 0.7783 on the deployment
decoder; against the teacher's own 20-draw re-sampling level it reaches $R=97.1\%$ with the recorded analysis and 92.3\%
with its own. The free-form channel is preserved and live: the generated analysis is close to the teacher's (cosine
0.943), swapping semantic inputs moves 64.7\% of decisions against 12.8\% for a matched donor, and a read-out of the
generated analysis recovers most of the decision signal. A five-seed, 2$\times$3 content-by-schedule study isolates what
carries the deployment gain: scheduled self-history does (+2.2 points over no-semantic retrains, with shuffled- and
constant-content controls reproducing the gain), while semantic content matters at the teacher-forced point where it
can be read. That is a practical rule for building low-compute semantic-preserving surrogates: spend the budget on the
deployment schedule and evaluate content at the teacher-forced operating point rather than assuming it must transfer.

\paragraph{A cheap instrument.}
Determinism is what makes the study affordable. With the simulator seed fixed, the surrogate re-runs exactly, so a
20-seed, 200{,}000-decision sweep takes 12 minutes and zero LLM calls, against about three hours and
$4.0\times10^5$ calls for the sampling teacher; replay diagnostics, seed replications and component controls fit in one
training budget. The first run still requires the teacher trace and one encoder pass, but every subsequent run removes
the teacher from the loop. The surrogate turns an expensive stochastic simulator into a cheap deterministic instrument
for repeated evaluation, seed sweeps and surrogate studies.

\paragraph{Scope and outlook.}
The method applies to staged LLM agents whose intermediate outputs are free-form text; the interface, visibility
matrix, joint stage losses and scheduled self-history are specified so the recipe can be transferred to a new trace. The
empirical study is a case study on one simulator, one recorded trace and one market regime, and the teacher reference is
a same-API re-sample of an unversioned teacher, so the fidelity numbers are relative agreement rather than an
equivalence claim. Within that scope the paper establishes the method and its transfer analysis; applying the recipe to
a second staged agent is the natural next step.

\paragraph{Conclusion.}
This paper proposed staged semantic distillation, a low-compute method for preserving free-form semantics in a non-LLM
surrogate of a staged LLM agent, and used the resulting deterministic student as an instrument to study what the
transfer carries. The surrogate keeps the analysis and rationale channel live, reaches 97.1\%/92.3\% of the teacher's
own re-sampling level, and identifies a transferable design rule: scheduled self-history carries the deployment gain,
while semantic content earns its value at the teacher-forced operating point. The method and the deterministic
instrument it produces make staged LLM-agent simulation cheaper to re-run and easier to study.

%% file: sections_v5/08_appendix.tex
\section{Appendix: Protocol, Metrics, and Implementation Details}
\label{sec:appendix}

\subsection{Data, simulator, and split}
The ten Chinese industry indices are simulated by the TwinMarket multi-agent market simulator: 20 LLM agents per index,
five decision rounds per day, 102 trading days (2023-06-15 to 2023-11-15), for 102{,}000 agent-decisions in total. An
order is executed against the simulated limit-order book subject to a $\pm10\%$ price band, 100-share lot rounding, and
cash/sellable-position constraints; hold decisions arise from the position and ordering rules rather than from a third
action class. Days are split 86/6/10 into train/validation/test, with 200 agents per day, i.e.\ 17{,}200 agent-day slots, per round-one split. The round-one test set
has 1{,}998 labelled decisions (2 slots without an executed order), the round-one train set has 17{,}058 of 17{,}200
(142 without an executed order), and the validation set has 1{,}197 of 1{,}200 (3 without). Labels are the executed side
of the teacher's order (buy/sell). The self-analysis operating point is evaluated on the 1{,}982 round-one rows that
lie inside the 1{,}982 complete five-round agent-days used by the autoregressive protocol: 18 of the 2{,}000 agent-days
are incomplete, and 16 of their round-one rows are labelled, which is why the self-analysis comparison uses 1{,}982
rows. The donor probe uses 1{,}995 rows (3 lack a donor pair). The one-step window has 9{,}956 labelled rows of 10{,}000,
and the autoregressive protocol uses 1{,}982 $\times$ 5 = 9{,}910 rows. The
interface analysis uses round one only; the autoregressive and
closed-loop evaluations use the full test window. The closed loop is initialised from the teacher's account state at the
start of the evaluation window and rolled over the 102 days; no future information is used.

Table~\ref{tab:rows} gives the row flow for every denominator used in the paper. The 1{,}982 self-analysis rows are the
round-one rows of the same 1{,}982 complete five-round agent-days used by the autoregressive protocol; the remaining 16
labelled round-one rows lie in the 18 incomplete agent-days that decoder excludes. The 46-row difference between the
one-step ($9{,}956$) and autoregressive ($9{,}910$) row sets is exactly the number of labelled rows inside those 18
agent-days; their other 44 slot rows are the labelled-missing rows, so the 18 agent-days account for all 90 excluded
slot rows.

\begin{table}[!htbp]
\centering
\footnotesize
\caption{Row flow for every denominator in the paper. ``Excluded'' gives the number of slot rows and the reason. The
1{,}982 self-analysis rows are the round-one rows of the 1{,}982 complete agent-days used by the autoregressive decoder,
so the two comparisons share that row set.}
\label{tab:rows}
\setlength{\tabcolsep}{3pt}
\begin{tabular}{lcc}
\toprule
Comparison & Rows & Excluded \\
\midrule
Round-one labelled decisions & 1{,}998 & 2 no executed order \\
Self-analysis & 1{,}982 & 16 incomplete agent-day \\
Donor probe & 1{,}995 & 3 no donor pair \\
One-step P1, full window & 9{,}956 & 44 no executed order \\
Autoregressive P2, full window & 9{,}910 & $18\times5$ slots, 46 labelled \\
Round-one train / val. & 17{,}058 / 1{,}197 & 142 / 3 no order \\
\bottomrule
\end{tabular}
\end{table}

\subsection{Teacher re-sampling protocol}
For each of the 1{,}998 round-one test contexts we replayed the teacher 20 times. The replay used the same API model as
the recorded trace (\texttt{deepseek-chat}, temperature 0.9, top-p 0.9, frequency and presence penalties 0.5, max
tokens 4096), i.e.\ the model and sampling settings of the simulator's analysis and decision calls (its reasoner model
is used only for the belief-update and forum stages, which are not replay targets). The API does not expose a model
snapshot identifier. The trace was generated with this model name on 2026-11-06 and the replay on 2026-11-10; the
provider exposes no immutable snapshot, so all parity and $R$ values are same-API re-sampling-relative quantities. To
measure version sensitivity, we re-replayed 128 contexts (every 15th) in both conditions on 2026-11-14: $S_A$ changes by
$-0.36$ points (95\% bootstrap interval $[-1.24,+0.44]$) and $S_B$ by $+0.26$ points ($[-1.73,+2.26]$), within replay
noise, so the denominator is stable over this interval; the check is stored as
\texttt{replay\_version\_check\_20261114.json}. All 39{,}960 condition-A and 79{,}920 condition-B replay requests
returned a response; the mean number of usable
parsed draws per context is 19.97 (A) and 19.98 (B), and we use the realised count. Condition A places
the recorded analysis text in the context and re-asks the decision prompt, so each draw generates a fresh rationale and
action; the recorded round rationale and action are not in the replayed prefix. Condition B regenerates the analysis
from the same context before each decision, so each draw costs two model calls. A
small number of contexts have fewer than 20 successful draws (mean 19.97 in A, 19.98 in B), and we use the realised
per-context count. The replayed decisions are evaluation-only: the trace is not modified and the student is never
trained on replay outputs.

\paragraph{Re-sampling level estimators.}
From the replay we estimate $\hat p_A$ and $\hat p_B$ as per-context action frequencies. The re-sampling level is the
expected agreement of a fresh draw with the recorded action, computed in balanced-accuracy space as
$\tfrac12(\hat p_{\text{buy}}(\text{recorded buy})+\hat p_{\text{sell}}(\text{recorded sell}))$ and in accuracy space as
the mean probability assigned to the recorded action; this is the empirical class-conditional form of Eq.~1; its balanced- and accuracy-space forms agree within 0.0003 on
these rows (0.9368 vs 0.9370 in A, 0.7860 vs 0.7863 in B). The finite 20-draw estimate $\hat p$ has its own binomial uncertainty. Our two-level bootstrap redraws
per-context replay draws from $\hat p$, so it captures within-context replay noise but conditions on $\hat p$ and does
not propagate the uncertainty of $\hat p$ itself; a hierarchical bootstrap that re-estimates $\hat p$ per replicate, or
100 draws on a subset, needs additional teacher calls and was not run under this study's constraint of no teacher calls
beyond the round-one replay. The
sample-fidelity intervals are therefore slightly optimistic, and $R$ should be read as a point diagnostic with an
interval that is a lower bound on its uncertainty.

\paragraph{Synthetic validation of the ratio.}
We checked the estimator and its interval on synthetic teachers with known ground truth. We draw 1{,}000 contexts with
$p\sim\mathrm{Beta}(5,5)$, a recorded action from $p$, and two deterministic students: a Bayes-optimal one in this
symmetric setting and one with 10\% of its predictions flipped. The target is the realised-sample expected-agreement
ratio, known because $p$ and the student are generated; we draw $K$ teacher replays per context, estimate $R$ with the
direct label-conditional estimator, and form the same two-level bootstrap interval (300 replicates, 200 outer replicates
per cell). Table~\ref{tab:synthetic} reports bias and coverage. At $K=20$ and $K=100$ the bias is at most 0.2 points of
$R$ and the 95\% interval covers the true ratio in 98.5--100\% of replicates, so the interval is conservative rather
than anti-conservative; at $K=5$ the bias stays below 0.6 points. This validates the $K=20$ replay used in the main
text at the level of the estimator. The simulation treats contexts as independent and uses 1{,}000 contexts, whereas
the real study has 1{,}998 contexts and ten date clusters; the two widths are therefore not comparable, and the
synthetic coverage is not a calibration of the real date-level interval. The date-level re-analysis in
Section~\ref{sec:interface} is the relevant uncertainty statement for the decision contrasts, and we do not use the
synthetic width to characterise the real one.

\begin{table}[!htbp]
\centering
\footnotesize
\caption{Synthetic-teacher validation of the sample-fidelity ratio. 1{,}000 contexts, $p\sim\mathrm{Beta}(5,5)$;
200 outer replicates per cell; two-level bootstrap with 300 replicates. Bias is the mean estimated minus true ratio,
in percentage points of $R$; the true ratio is about 1.37 for the Bayes student and 1.31 for the flipped one.
Coverage is the fraction of 95\% intervals containing the true ratio.}
\label{tab:synthetic}
\setlength{\tabcolsep}{4pt}
\begin{tabular}{lcccc}
\toprule
Student & $K$ & Bias (pts of $R$) & 95\% CI width & Coverage \\
\midrule
Bayes-optimal & 5 & $-0.01$ & 0.205 & 1.000 \\
 & 20 & $-0.15$ & 0.185 & 1.000 \\
 & 100 & $-0.19$ & 0.176 & 1.000 \\
Flipped 10\% & 5 & $-0.52$ & 0.197 & 0.985 \\
 & 20 & $-0.06$ & 0.176 & 1.000 \\
 & 100 & $+0.02$ & 0.171 & 0.995 \\
\bottomrule
\end{tabular}
\end{table}

\subsection{Stage visibility matrix}
\label{sec:mask}
Table~\ref{tab:mask} gives the attention mask of the deployed student exactly as implemented. Each stage is an
attention prefix over the token layout shown; a filled cell means the stage's query can attend to that token group. The
analysis stage has no access to the current day's rounds or semantic history, the rationale and action stages see the
full context, and the action head additionally reads the generated analysis and rationale vectors directly (the
concatenation in the model's action read-out) in addition to attending to their tokens. The news query is an auxiliary
head: retrieved news is precomputed and fixed, and the predicted query does not drive retrieval in this implementation.
The indicator query and summary are generated in-pass and consumed by the later stages.

\begin{table*}[t]
\centering
\footnotesize
\caption{Stage visibility matrix of the deployed student. Columns: P persona, N retrieved news (two tokens), Q indicator
query, V indicator values, S indicator summary, A day analysis, C account state and extra context, H previous-round
parameters, previous actions, competition and day-cumulative features, R current-day round history, M semantic history
(analysis and rationale of earlier rounds), D previous-day history, T current rationale. Filled cells are the token
groups each stage's attention prefix admits.}
\label{tab:mask}
\setlength{\tabcolsep}{3.2pt}
\begin{tabular}{lcccccccccccc}
\toprule
Stage & P & N & Q & V & S & A & C & H & R & M & D & T \\
\midrule
News query & $\bullet$ & & & & & & & & & & & \\
Indicator query & $\bullet$ & & & & & & & & & & $\bullet$ & \\
Indicator summary & $\bullet$ & & $\bullet$ & $\bullet$ & & & & & & & & \\
Day analysis & $\bullet$ & $\bullet$ & & & $\bullet$ & & $\bullet$ & & & & $\bullet$ & \\
Rationale & $\bullet$ & $\bullet$ & & & $\bullet$ & $\bullet$ & $\bullet$ & $\bullet$ & $\bullet$ & $\bullet$ & $\bullet$ & \\
Action & $\bullet$ & $\bullet$ & & & $\bullet$ & $\bullet$ & $\bullet$ & $\bullet$ & $\bullet$ & $\bullet$ & $\bullet$ & $\bullet$ \\
\bottomrule
\end{tabular}
\end{table*}

\subsection{No-semantic-history retrain}
The retrain uses the identical architecture (three layers, width 384, 13{,}175{,}439 parameters), data,
split, optimizer and early-stopping rule as the deployed student, with the semantic-history tensor and its mask zeroed
in both training and evaluation; because the semantic-history slots are absent, the scheduled self-history over those
slots is not applied, and this is a property of the control rather than a separately matched schedule. The semantic
output losses remain on, so the model still emits stage vectors and its action head still reads the generated analysis
and rationale vectors, but no analysis, rationale or other semantic token from the history enters its attention
context. It therefore measures the semantic-history package (slots, masks and schedule), not the in-pass stage chain. It early-stopped at epoch 14 with best validation NLL 0.4157
(deployed: 0.3641) and was evaluated with the same P1 and P2 protocols with the mask applied at test time, so no
test-time distribution shift is involved. Five seeds of this retrain reach 0.7842/0.7805/0.7919/0.7896/0.7778 one-step
(mean 0.7848, sd 0.60) and 0.7498/0.7561/0.7659/0.7535/0.7508 autoregressive (mean 0.7552, sd 0.65), with validation
NLL 0.4157/0.4178/0.4227/0.4241/0.4203. The paired date$\times$asset differences from the deployed checkpoint
(seed 0) against the five retrain seeds are $+5.0$/$+5.4$/$+4.3$/$+4.5$/$+5.7$ one-step and
$+2.9$/$+2.2$/$+1.2$/$+2.5$/$+2.8$ autoregressive, every interval excluding zero; against the retrain mean the
deployed checkpoint contributes $+5.0$ (teacher-forced, the analysis-input + channel package) and $+2.3$
(autoregressive, the semantic-history package; the three-seed shuffled-content control in Appendix~\ref{sec:shuf} shows that this gain does not require agent-matched content). Across the five deployed-recipe seeds the arm-mean
gains are $+4.6$ [3.6,5.6] and $+2.2$ [1.21,3.14] (two-level bootstrap over training seeds and date$\times$asset
clusters), and every deployed seed's autoregressive value is above every retrain seed's. The retrain was trained five
times and the deployed recipe five times. The higher validation NLL of the
retrains is not evidence that they are undertrained in the relevant sense: each checkpoint is its best validation
epoch, so early stopping is applied identically, and the absent semantic history is what makes the next decision harder
to predict; we cannot separate that from a general training-quality difference, which is why we report the direction
and the seed-aware interval rather than a fidelity statement.

\subsection{Within-date shuffled-content control}
\label{sec:shuf}
The control uses the same architecture, data, split, optimizer and early-stopping rule as the deployed student, but
with a content permutation: every recorded analysis/rationale vector in the semantic-history tensor is replaced by that
of another agent on the same date and round, separately within each split. Masks, schedule and the date-specific action
composition are unchanged, so the vector distribution and the exposure protocol are matched while the agent-to-vector
identity link is broken. Three training seeds were run (Table~\ref{tab:controls}); their full-test one-step values are
0.7818/0.7874/0.7879 (mean 0.7857), autoregressive 0.7816/0.7847/0.7801 (mean
0.7821) and round-one 0.7254/0.7483/0.7347 (mean 0.7361). The
round-one seed range is 2.3
points, so the control does not support a single-seed reading of the round-one deficit. Against the deployed arm the
arm-mean differences are -4.54 points one-step,
+0.45 autoregressive and +0.22
at round one; against the no-semantic retrain the autoregressive difference is
+2.69 and the round-one difference is
-1.61. The loader re-applies the same within-split shuffle at
evaluation (the flag is stored in the checkpoint), so each control is tested on the history distribution it was trained
on; its validation NLL is computed on the shuffled validation split and is not comparable with the deployed model's
0.3641 on recorded history, since both are early-stopping criteria on their own information set.

\subsection{Constant-content channel control}
\label{sec:contentfree}
The control uses the same architecture, data, split, optimizer and early-stopping rule as the deployed student, but
keeps the semantic-history slots, their masks and the full round schedule while setting every value in the tensor to the
training mean (zero in normalised space) in both training and evaluation. The scheduled content substitutions are
applied and then zeroed, so the semantic schedule machinery runs but contributes no content; the action-history
schedule and the current-analysis input are unchanged. It therefore removes the history content while holding the
current-analysis input fixed. The semantic schedule is effectively inert in this arm, which is why Section~\ref{sec:interface}
adds no-schedule cells at each content level. The no-schedule cells use the same architecture,
data, split, optimizer and early-stopping rule with all three scheduled-sampling probabilities set to zero. Full-test
one-step/autoregressive values are: shuffled 0.7907/0.7591, constant 0.7841/0.7534, to be read against the
schedule-on cells 0.7818/0.7816 and 0.7941/0.7857 and the matched-content cells 0.8347/0.7783 (on) and 0.8143/0.7193
(off). The schedule effects on the autoregressive decoder are +5.90 [4.47,7.23] (matched), +2.25 [1.23,3.40]
(shuffled) and +3.22 [2.16,4.26] (constant), all paired date-level intervals excluding zero; with the schedule on,
content changes the decoder by at most +0.74 [-0.07,1.70], and the matched-versus-control difference-in-differences is
+3.65 [1.70,5.68] (versus shuffled) and +2.68 [1.09,4.16] (versus constant). The cells are one training seed each;
\texttt{round16\_nosched.py} and \texttt{round16\_nosched.json} regenerate the table. Three training seeds were run (Table~\ref{tab:controls}); their one-step values are
0.7941/0.7909/0.7891 (mean 0.7914), autoregressive
0.7857/0.7857/0.7795 (mean 0.7836) and round-one
0.7454/0.7538/0.7475 (mean 0.7489). Against the deployed arm the arm-mean
differences are -3.97 one-step,
+0.60 autoregressive and
+1.50 at round one; the autoregressive difference is within the
pre-specified 1.5-point equivalence margin (TOST p = 0.012),
and the round-one values overlap the deployed seed range. The control separates a present-but-content-free interface
from one carrying another agent's content; at the teacher-forced point the content matters, at the deployment decoder
it does not.

\begin{table}[!htbp]
\centering
\footnotesize
\caption{Seed-level content controls. P1 = full-test teacher-forced one-step balanced accuracy (9{,}956 rows); P2 = full-test within-day autoregressive balanced accuracy (9{,}910 rows); r1 = round-one self-analysis balanced accuracy (1{,}982 rows). The deployed and no-semantic rows give the five-seed mean; the control rows give individual training seeds and their three-seed mean. The shuffled control uses another agent's recorded same-date/round analysis/rationale with a fixed within-split permutation; the constant-content control sets every semantic-history value to the training mean and is evaluated content-free.}
\label{tab:controls}
\setlength{\tabcolsep}{3.5pt}
\begin{tabular}{llccc}
\toprule
Arm & Seed & P1 & P2 & r1 \\
\midrule
Deployed & 0--4 mean & 0.8311 & 0.7776 & 0.7339 \\
No-semantic & 0--4 mean & 0.7848 & 0.7552 & 0.7522 \\
Shuffled & seed 0 & 0.7818 & 0.7816 & 0.7254 \\
Shuffled & seed 1 & 0.7874 & 0.7847 & 0.7483 \\
Shuffled & seed 2 & 0.7879 & 0.7801 & 0.7347 \\
Shuffled & mean & 0.7857 & 0.7821 & 0.7361 \\
Constant-content & seed 0 & 0.7941 & 0.7857 & 0.7454 \\
Constant-content & seed 1 & 0.7909 & 0.7857 & 0.7538 \\
Constant-content & seed 2 & 0.7891 & 0.7795 & 0.7475 \\
Constant-content & mean & 0.7914 & 0.7836 & 0.7489 \\
\bottomrule
\end{tabular}
\end{table}

\subsection{Sample-fidelity ratio and distributional metrics}
Sample fidelity is reported as $R=\text{student balanced accuracy}/\text{teacher re-sampling level}$, both computed on
the same recorded decisions. The deployed checkpoint gives $R=97.1\%$ [95.8, 98.4] in condition A and $R=92.3\%$
[89.0, 95.4] in condition B; the interface checkpoint's seed 0 gives $R=98.7\%$ [97.4, 99.8] and $R=95.8\%$ [93.2, 98.2].
Ratios are computed with a two-level bootstrap: date$\times$asset clusters (100 clusters, 1{,}000 replicates, seed 0) are resampled and the per-context replay draws are redrawn from $\hat p$ within each replicate, so the intervals include
both context and replay-sampling noise.

\paragraph{Behavioural ratio versus replay-Bayes level.}
The ratio $R$ is defined against the teacher's own re-sampling agreement, not against the best deterministic predictor of the
replayed conditional. The replay-Bayes level is the accuracy of the plug-in argmax of $\hat p$ against the recorded
actions; on the same 20 draws that estimate the argmax, it is in-sample (0.9516 in condition A and 0.8406 in B). The
stored replay dumps retain the per-context fractions but not the draw sequence, so we compute the out-of-sample
analogue analytically: estimate the argmax from 10 draws and score it against the recorded action, integrating over the
binomial split under $\hat p$. That gives 0.9480 in A and 0.8302 in B, i.e.\ the in-sample values are optimistic by 0.4
and 1.0 points; the deployed student reaches 96.0\% and 87.4\% of the split-half level. The reference used in the main
text is the direct label-conditional estimator (0.9368 on the 1{,}998 teacher-analysis rows; 0.7858 on the matched
1{,}982 self-analysis rows). A quadratic plug-in of the same expected-agreement quantity, with the class frequencies
aligned to the class coding and the binomial bias subtracted, gives 0.9384 / 0.7902 (raw: 0.9414 / 0.8004; bias:
0.0030 / 0.0101), so the two estimators agree within 0.16 and 0.44 points after the correction and $R$ is
97.0--97.1\% and 91.8--92.3\% under either. $R$ is a diagnostic ratio, not a ceiling: a deterministic mode-seeker can
exceed $R=1$. A constant predictor, the chance-level deterministic policy, attains $R=53.4\%$ in condition A and
$63.6\%$ in condition B.

Distributional metrics compare the student side probability $q_{\text{buy}}$ with the replayed teacher probability
$\hat p_{\text{buy}}$ on the same rows: binary Brier score, total variation, KL divergence, expected calibration error
in ten equal-mass bins against $\hat p$, the fraction of contexts whose Argmax action agrees with the re-sampled
teacher mode, and the NLL of the recorded action under $q$. The finite replay count adds binomial noise to $\hat p$;
this inflates the plug-in Brier and total variation slightly and attenuates the KL, and the reported two-level
bootstrap intervals include this source of noise.

\begin{table}[!htbp]
\centering
\footnotesize
\caption{Distributional fidelity of the student's side probability to the re-sampled teacher conditional (round-one
test; 20 replays per context). Brier is binary; TV is total variation; ECE is ten-bin expected calibration error
against $\hat p$; NLL is the negative log-likelihood of the recorded action; mode is the fraction of contexts with the
same most likely action as the re-sampled teacher. Date$\times$asset cluster-bootstrap 95\% intervals for the deployed
rows, teacher-analysis / self-analysis: Brier [0.028, 0.040] / [0.072, 0.094]; TV [0.079, 0.097] / [0.202, 0.237];
KL [0.098, 0.135] / [0.209, 0.262]; ECE [0.010, 0.025] / [0.047, 0.093]; NLL [0.181, 0.234] / [0.516, 0.589]; mode
[0.918, 0.943] / [0.735, 0.803]. A constant base-rate forecast attains Brier 0.219, TV 0.454, KL 0.600 and mode 0.554
on the teacher-analysis rows.}
\label{tab:distribution}
\setlength{\tabcolsep}{2.5pt}
\begin{tabular}{lcccccc}
\toprule
Operating point & Brier & TV & KL & ECE & NLL & Mode \\
\midrule
Deployed, teacher analysis & \textbf{0.033} & 0.088 & 0.116 & \textbf{0.013} & 0.206 & 0.931 \\
Deployed, self analysis & 0.082 & 0.219 & 0.234 & 0.069 & 0.552 & 0.770 \\
Interface, teacher analysis & 0.027 & 0.086 & 0.098 & 0.019 & 0.189 & 0.945 \\
\bottomrule
\end{tabular}
\end{table}

\subsection{Design study}
All design-study variants share the data, split, and evaluation protocol. The schedule variants differ in the training
recipe: removing self-history scheduling (all probabilities set to 0); the action-only behaviour-cloning control
(analysis-only schedule, all five semantic stage losses set to 0); the matched analysis-only schedule ($p_{\mathrm{self}}=0.5$ on
the analysis slot only, stage losses on); the full schedule with all five semantic stage losses set to 0; the full schedule with
the stage losses on ($0.5$ analysis + $0.35$ rationale and action-history slots); and the no-semantic-history retrain
(semantic tokens masked in training and evaluation, stage losses kept). The four cells analysis-only on/off and full
on/off give the 2$\times$2 reported in Section~\ref{sec:interface}: 0.8218/0.7419, 0.8226/0.7324, 0.8265/0.7660 and
0.8347/0.7783. The
structural variant uses the same capacity and full schedule but ties the semantic output heads to the input projections
(the output map is the transpose of the input projection) and removes the action bypass (the action head reads only the
semantic vectors, not the full action representation). Full-test metrics: one-step balanced accuracy on the P1
teacher-forced replay ($n=9{,}956$) and within-day autoregressive balanced accuracy on the P2 rollout ($n=9{,}910$). The
analysis-only, tied+no-bypass, and full-schedule variants were run in closed loop; the two schedule ablations were not,
and the macro column of Table~\ref{tab:design} reports the three available values.

The six-day validation NLL (temperature 1, computed from the saved validation logits; the split is not part of the test
window) is 0.3641 for the deployed checkpoint (seeds 0.3641/0.3562/0.3585/0.3605/0.3519), 0.3660 for the action-only control, 0.3700 for the analysis-only schedule
(0.3700, 0.3681, 0.3700 across its three seeds), 0.3679 for the no-schedule variant, 0.3465 for the tied+no-bypass
model, 0.3577 and 0.3642 for the 192- and 256-width students, and 0.4157 for the no-semantic retrain. Selecting the
variant on this held-out criterion would choose the tied+no-bypass model, not the deployed full-schedule model; that
checkpoint is statistically tied with the deployed model on both test decoders (Section~\ref{sec:interface}), so the
decision-level conclusions are insensitive to the choice, while the macro axis that prefers the deployed model is
in-sample.

The tied+no-bypass variant has the best round-one sample fidelity of the family (97.9\% [96.7, 99.2] and 93.4\%
[90.6, 96.0]) and the best unweighted test accuracy (0.8415 / 0.7878), but its balanced accuracy is 0.15/0.26 points
lower (0.8332 / 0.7757), within the paired date$\times$asset interval. The deployed model keeps independent stage heads
and the full action representation because it is best on both decoders or tied within the paired interval, and its
closed-loop statistics were the closest of the three variants run in closed loop, although that axis participated in
the selection: on the ten-index set 0.68 and 156.9 bp against the
ten-index teacher path's 0.90 and 191.0 bp, and on the six-index set 0.38 and 162.1 bp against 0.66 and 168.3 bp.
The macro axis participated in the selection, so these closed-loop comparisons are descriptive rather than held-out
validation.

\begin{table}[!htbp]
\centering
\footnotesize
\caption{Design study on the full test window (one-step $n=9{,}956$; within-day autoregressive $n=9{,}910$). All
variants share the data, split and evaluation protocol; balanced accuracy is reported. The last column gives the
closed-loop tail level for the variants that were run in closed loop (Section~\ref{sec:deployment}); those rows are
ten-index, and the ten-index teacher value is 0.90 (the six-index teacher value is 0.66). The deployed model is the
full-schedule variant with independent stage heads and the full action representation. The action-only row is the
behaviour-cloning control: all five semantic stage losses are set to zero and only the analysis slot is scheduled. With the
full-schedule/no-stage-loss row it completes a 2$\times$2 over schedule extent (analysis-only versus full) and stage
supervision (on/off), reported in the text; the cells are one training seed each. The no-semantic row is seed 0 of a
five-seed retrain with the semantic-history input masked during training and evaluation
(Section~\ref{sec:interface}); its five-seed full-test values are 0.7842/0.7805/0.7919/0.7896/0.7778 and
0.7498/0.7561/0.7659/0.7535/0.7508. The shuffled-content row keeps the same slots, masks and schedule but takes every recorded
analysis/rationale vector from another agent on the same date and round (three training seeds), so it changes the
agent-to-vector identity link while holding the vector distribution and the date-specific action composition fixed. The content-free row keeps the slots, masks and schedule but sets every
semantic-history value to the training mean (zero in normalised space) and is evaluated content-free (three training seeds;
Appendix~\ref{sec:contentfree}); per-seed values are in Table~\ref{tab:controls}. The two no-schedule content rows
repeat the shuffled and constant-content cells with all three scheduled-sampling probabilities set to zero (one training
seed each); with the matched-content/no-schedule row they form the 2$\times$3 content-by-schedule design of
Table~\ref{tab:contentsched}.
The flat-token row has the
same input features and a matched parameter budget but no stage tags, visibility, stage heads or stage losses, so it is
one-step only (it cannot generate stage outputs). The flat MLP is an action-only behaviour-cloning model on the numeric
context with no semantic channel; the HGB row is the state-only structured-feature baseline, also one-step.
The flat-token model sees the recorded semantic inputs while the no-semantic retrain does not, so their gap (3.35
points) is an information-set and depth difference, not a typing effect; the typing-only contrast is the flat-token row
against the staged deployed row at the same inputs (1.70 points).}
\label{tab:design}
\setlength{\tabcolsep}{1.4pt}
\begin{tabular}{lccc}
\toprule
Variant & One-step bal. & AR bal. & Closed-loop kurt. \\
\midrule
No self-history schedule & 0.8143 & 0.7193 & -- \\
Shuffled content, no schedule & 0.7907 & 0.7591 & -- \\
Constant content, no schedule & 0.7841 & 0.7534 & -- \\
Action-only BC (analysis sch.) & 0.8226 & 0.7324 & -- \\
Analysis-only schedule & 0.8218 & 0.7419 & 2.30 $\pm$ 0.58 \\
Full sched., no stage losses & 0.8265 & 0.7660 & -- \\
No semantic hist. (retrain s0) & 0.7842 & 0.7498 & -- \\
Content-free channel (3 seeds) & 0.7914 & 0.7836 & -- \\
Shuffled semantic content (3 seeds) & 0.7857 & 0.7821 & -- \\
Tied + no-act-bypass & 0.8332 & 0.7757 & 2.42 $\pm$ 0.38 \\
Flat-token transformer (no typing) & 0.8177 & -- & -- \\
Flat MLP on numeric context & 0.7779 & -- & -- \\
HGB on structured state & 0.8057 & -- & -- \\
Full schedule (deployed) & \textbf{0.8347} & \textbf{0.7783} & \textbf{0.68 $\pm$ 0.11} \\
\bottomrule
\end{tabular}
\end{table}

\subsection{Per-round accuracy and the seed rule}
Table~\ref{tab:perround} gives the per-round balanced accuracy of the deployed checkpoint, the no-semantic retrain and
the second training seed on both decoders. The deployed model is above the retrain on every round 2--5 of the
autoregressive decoder (+6.9, +1.6, +5.3 and +3.3 points) and below it at round one ($-2.3$); the same pattern holds for
the second seed. The one-step decoder shows the round-one teacher-forced contrast (+16.3 points) and smaller gains in
rounds 2--5 (+3.2, +1.4, +4.2, +0.9). This is the table form of the claim that the channel's autoregressive gain is
carried by rounds 2--5.

\begin{table}[!htbp]
\centering
\footnotesize
\caption{Per-round balanced accuracy ($n=1{,}982$ per round on the autoregressive decoder; 1{,}989--1{,}998 on the
one-step decoder). The retrain never receives semantic history; it is insensitive to the analysis slot by construction,
so its one-step and self-analysis numbers are the same baseline. Per-round entries are unweighted means of round-specific balanced accuracies; the pooled full-test values reweight rounds by class frequency, which is why the pooled value differs from the row mean by 0.14--0.19 points for the retrain and the second seed and by 0.03 (autoregressive) or 0.006 (one-step) for the deployed model.}
\label{tab:perround}
\setlength{\tabcolsep}{3pt}
\begin{tabular}{lccccc}
\toprule
Decoder / model & R1 & R2 & R3 & R4 & R5 \\
\midrule
One-step, deployed & \textbf{0.9098} & 0.8456 & 0.7949 & 0.8222 & 0.8007 \\
One-step, retrain & 0.7470 & 0.8137 & 0.7807 & 0.7807 & 0.7914 \\
One-step, second seed & 0.9137 & 0.8289 & 0.7753 & 0.7920 & 0.8189 \\
Autoregressive, deployed & 0.7252 & \textbf{0.8196} & \textbf{0.7752} & \textbf{0.7961} & 0.7740 \\
Autoregressive, retrain & 0.7485 & 0.7502 & 0.7588 & 0.7436 & 0.7407 \\
Autoregressive, second seed & 0.7348 & 0.8110 & 0.7660 & 0.7779 & \textbf{0.7866} \\
\bottomrule
\end{tabular}
\end{table}

\paragraph{Seed-selection rule.}
The deployed checkpoint (\texttt{multimodal\_5r\_hero\_s0\_best.pt}, training seed 0) was the first run of the deployed
recipe and was fixed for all headline results, full-window and closed-loop evaluations before any further seed was
trained. The four further seeds (training seeds 1--4) are replications only; they were trained after the headline and
closed-loop results were fixed and are not used to select any checkpoint, hyperparameter or variant. No test-window
quantity was used in the seed choice. The checkpoint epoch within each run is selected on the six-day validation NLL,
and the choice among design variants used the in-sample macro axis, as stated in Section~\ref{sec:interface}.

\paragraph{Semantic removal probe.}
The probe reloads the deployed checkpoint and re-runs the full-test teacher-forced rows with the semantic-history tensor
altered (tokens masked; token values set to the training mean, i.e.\ zero in normalised space; N(0,1) values; or the
analysis/rationale half of each active slot set to its mean separately); the
checkpoint, data, and masks are otherwise unchanged. The unchanged pass gives 0.8344 balanced accuracy (0.8347 in the
main P1 pass, a different batch pass over the same rows). No model is retrained for this probe and no teacher call is
made; the retrained no-semantic-history student above is a separate model and measures the value of the interface rather
than the trained student's reliance on it.

\subsection{Closed-loop macro protocol}
The closed loop is run for the full 102-day window on the ten indices, with the deployed checkpoint, teacher
initialisation, and three simulator seeds per decoding rule ($1$--$3$); only the simulator seed changes between runs.
Deterministic decoding uses the model's Argmax side action and its median size/offset; sampled decoding draws from the
model's predicted distributions. Daily simple returns are computed from the daily close series; excess kurtosis is
$\mathbb{E}[(r-\mu)^4]/\sigma^4-3$ (moment estimator), volatility is the standard deviation in basis points,
ARCH(1) is the lag-1 autocorrelation of squared daily returns, and the maximum daily move is the largest absolute daily
return. Per-seed statistics are means over the ten indices; we report the mean $\pm$ sd across the three seeds.

For the teacher macro reference we also report the six indices whose recorded path stays within 40\% of the real
index's end-of-window level and within 60\% of its volatility (CPEI, EREI, FSEI, MEI, TSEI, TTEI); the remaining four
teacher paths show a strong end-of-window level drift. This six-index subset is selected on the teacher's own closeness
to the real index and is therefore outcome-dependent and secondary; the ten-index rows are the primary macro check. The real-market row is computed on the ten indices over the same window.

\begin{table}[t]
\centering
\footnotesize
\caption{Closed-loop macro statistics over the 102-day window. Rows marked (6) use the six indices of the teacher
reference and rows marked (10) are means over ten indices; student rows additionally show the number of simulator seeds
and report mean $\pm$ sd across seeds (each seed is the mean over its index set). Kurtosis and volatility show the
seed mean $\pm$ seed sd; maximum move and ARCH(1) are seed means. The teacher path is one recorded
realization, not a distribution over teacher runs. The same checkpoint, initialisation and window are used across both
decoding rules; the decoding rule and the simulator seed change across rows, and within a decoding rule only the seed
changes. $^{\dagger}$Teacher recorded path: mean over the six indices whose
end-of-window level stays within 40\% of the real index and whose volatility stays within 60\% of it (CPEI, EREI, FSEI,
MEI, TSEI, TTEI); the other four paths show a strong end-of-window level drift. The all-ten rows are primary and the six-index rows secondary/outcome-selected. The analysis-only and tied+no-bypass
entries are design variants. $^{\ddagger}$The real-market maximum (4.727\%) and the deployed six-index maximum
(4.727\%) coincide to three decimals because both are means of per-index maxima over different index sets: the
real-market figure is the ten-index mean (4.7273\%), while the deployed six-index figure is the mean over its three
seeds of the six-index means (4.6365/4.7723/4.7714, i.e.\ 4.7267\%); they are different series. The ten-index values are
5.97\% (teacher) and 4.766\% (deployed). Volatility in basis points; ARCH(1) is the lag-1 autocorrelation of squared daily returns.}
\label{tab:macro}
\setlength{\tabcolsep}{1.9pt}
\begin{tabular}{lcccc}
\toprule
Series & Excess kurt. & Max move & Vol (bp) & ARCH(1) \\
\midrule
Real market (10) & 1.66 & 4.727\% & 137.2 & $-0.05$ \\
Teacher path$^{\dagger}$ (6) & 0.66 & 5.15\% & 168.3 & 0.13 \\
Teacher path (10) & 0.90 & 5.97\% & 191.0 & 0.13 \\
Deployed det.\ (6, 3) & 0.38 $\pm$ 0.18 & 4.727\% & 162.1 & 0.06 \\
Deployed samp.\ (6, 3) & 0.03 $\pm$ 0.30 & 3.87\% & 139.7 & 0.04 \\
Deployed det.\ (10, 3) & \textbf{0.68 $\pm$ 0.11} & \textbf{4.766\%} & \textbf{156.9} & \textbf{0.07} \\
Deployed samp.\ (10, 3) & 0.12 $\pm$ 0.25 & 4.00\% & 140.0 & 0.07 \\
Analysis-only variant (10, 3) & 2.30 $\pm$ 0.58 & 3.83\% & 109.2 & 0.08 \\
Tied+no-bypass variant (10, 3) & 2.42 $\pm$ 0.38 & 4.60\% & 130.5 & 0.02 \\
\bottomrule
\end{tabular}
\end{table}

\subsection{Fair baselines}
The fair baseline is gradient-boosted trees (HGB) trained on the student's structured input features: state (71),
extra (22), previous-round prompt parameters (19), previous-round signed-action features (7), competition (4),
day-cumulative (4), indicator values (55), per-round history ($5\times18$), day history ($10\times12$), asset identity,
PCA-64 of the retrieved news vectors with their mask, and PCA-64 of the persona vector. The with-teacher-analysis
baseline adds the teacher's analysis vector; the analysis-only reader reads the student's own generated analysis
vector. No teacher indicator summary or other oracle input is used. HGB uses 300 iterations, learning rate 0.06, 31
leaves and L2 regularisation 1.0, with early stopping on an internal 12\% validation split of the training rows; these
are the same fixed settings across the reader family and were set before the test rows were read, so no reader
hyperparameter was selected on the test window. Three reader seeds give a standard deviation of about 0.1 points. The
analysis-only reader reaches 0.7516 [0.721, 0.777] HGB / 0.7567 [0.730, 0.779] logistic on
the deployed checkpoint's generated vectors (one seed each); the family's three-seed sd is about 0.4 points, and the
paper labels which checkpoint a reader value belongs to. The teacher-analysis text-versus-vector control is a linear read-out trained and calibrated
on the same rows.

\paragraph{Full-window baseline.}
For the full-window comparison of Section~\ref{sec:interface} we train the same HGB family on all labelled train rows
(r1--r5) and evaluate it on the 9{,}956 labelled test rows. The strongest variant, which is given the teacher's
recorded analysis vector in addition to the structured features, reaches 0.8427 accuracy / 0.8343 balanced accuracy;
the state-only variant reaches 0.8160 / 0.8057. The deployed student reaches 0.8358 / 0.8347 in the same
teacher-forced condition, where both the reader and the student receive the recorded analysis vector as input; it
therefore matches the with-teacher-analysis reader on balanced accuracy, although the reader consumes recorded structured
features while the student generates its intermediate stages, so this is not a full information-set match. In the
self-analysis operating point the student does not receive the teacher's analysis; the corresponding read-out
comparison is reported in Section~\ref{sec:interface}. The reader is a one-step predictor that consumes recorded inputs; it does not
generate the unstructured semantic stages the student produces. Artifacts: \texttt{hgb\_r1r5.py},
\texttt{hgb\_r1r5.json}, \texttt{hgb\_r1r5\_nohist.py}, \texttt{hgb\_r1r5\_nohist.json}.

\subsection{Self-consistency protocol}
The round-one protocol has two forward passes. The first clears the current-analysis slot and produces the no-analysis
action. The second injects the student's own generated analysis into the slot and produces the read-back action.
Self-consistency is the balanced agreement between the two actions over the 1{,}998 round-one test decisions; the
deployed checkpoint gives 0.9325 and the interface checkpoint 0.9121.

\begin{table}[!htbp]
\centering
\footnotesize
\caption{Encoding read-out, head re-training and self-consistency (round one). Reader rows: HGB or logistic on the
student's state features (checkpoint-independent), on the deployed checkpoint's own generated analysis vector, and on
the frozen-backbone head features $[\text{act};\mathbf{a};\mathbf{r}]$ with a head re-trained on the 17{,}058 training
rows and evaluated on the full 1{,}998 test rows; the tabulated head values are on the matched 1{,}982 rows
(256-unit MLP 0.7468 $\pm$ 0.0106, 4,866-parameter linear refit 0.7425 $\pm$ 0.0061, 4,872-parameter MLP
0.7483 $\pm$ 0.0123, 2,433-parameter logistic 0.7439; the 4,872-parameter MLP has two hidden units, i.e.\ $2{,}432\times2+2=4{,}866$
parameters into the hidden layer plus $2\times2+2=6$ into the output layer). The family's three-seed reader sd is about 0.4 points. All reader values are balanced accuracy with 95\% date$\times$asset cluster-bootstrap intervals
where available. Self-consistency: two own passes, action with an empty analysis slot versus action after injecting the
student's own analysis, on 1{,}998 round-one rows.}
\label{tab:readers}
\setlength{\tabcolsep}{3pt}
\begin{tabular}{lc}
\toprule
Read-out & Balanced accuracy \\
\midrule
State only (HGB) & 0.7424 [0.713, 0.770] \\
Analysis only (HGB, deployed H) & 0.7516 [0.721, 0.777] \\
Analysis only (logistic, deployed H) & 0.7567 [0.730, 0.779] \\
Teacher analysis only (HGB) & \textbf{0.9076} [0.894, 0.921] \\
Re-trained head, 4,866-param linear (3 seeds) & 0.7425 $\pm$ 0.0061 \\
Re-trained head, 2,433-param logistic & 0.7439 [0.717, 0.768] \\
Re-trained head, 4,872-param MLP (3 seeds) & 0.7483 $\pm$ 0.0123 \\
Re-trained head, 256-unit MLP (3 seeds) & 0.7468 $\pm$ 0.0106 \\
\midrule
Student self-consistency (deployed; 1{,}998 rows) & \textbf{0.9325} \\
Student self-consistency (interface; 1{,}998 rows) & 0.9121 \\
\bottomrule
\end{tabular}
\end{table}

\paragraph{Parameter-matched head refits.}
The deployed action head is a 4{,}866-parameter linear map from the 2{,}432-d vector to two softmax logits. To separate
capacity from objective we refit that exact architecture on the same frozen features and round-one rows (three seeds,
0.7425 $\pm$ 0.0061 on the matched 1{,}982 rows; per-seed gains $+1.18$/$+1.49$/$+2.36$ points, per-seed paired
intervals [$-0.7$,$+3.0$], [$-0.4$,$+3.4$], [$+0.6$,$+4.1$], three-seed $t$ interval [0.16,3.20]), a
4{,}872-parameter two-hidden-unit MLP (0.7483 $\pm$ 0.0123; gains $+3.21$/$+2.70$/$+0.88$, $t$ interval
[$-0.8$,$+5.3$]), and a 2{,}433-parameter one-logit logistic head (0.7439; $+1.82$, paired interval
[$-0.2$,$+3.9$]). The fresh heads are trained on round-one rows only, while the deployed head sees all rounds with the
joint objective, so the comparison changes capacity, training rows and objective together; the matched-capacity result
shows that the recovery is not a capacity artifact on this backbone, but it does not isolate the objective or rule out
round-one specialisation. To test whether the seed-0 recovery is checkpoint-specific we repeated the refit on all five
deployed backbones (two head seeds each, same rows, features and architecture). The per-backbone gains are
$+1.7$/$+0.1$/$+3.1$/$-1.6$/$-0.4$ points; the mean is $+0.6$ (sd 1.9, $t$ interval [$-1.7$,$+2.9$]), which includes
zero, and the sign changes across backbones. The seed-0 recovery is therefore a checkpoint-level point estimate, not
evidence of a recipe-level head deficit. Artifacts: \texttt{round12\_head.json},
\texttt{round15\_head\_backbones.json}.

\subsection{Additional consistency checks}
\paragraph{Teacher-analysis read-outs.}
Table~\ref{tab:teacherread} collects the four teacher-analysis read-outs used in the paper. They come from separate
control runs with different readers and feature sets and are not recomputed under one protocol; the table states what
each is and which comparison uses it. The HGB on the teacher's recorded analysis vector (0.9076) is the primary baseline
for the 15.6-point content gap; the structured-plus-analysis HGB (0.9044) is the full-information-set fair baseline; the
matched logistic read-outs of the analysis text (0.9205) and of its 1024-d vector (0.9180) are the frozen-encoder
control, which asks what the encoder costs relative to the analysis text. The 1.04-point HGB-versus-logistic vector
difference is a reader-family difference, in the same direction as on the generated vectors (0.7567 logistic vs.\
0.7516 HGB), and the two were not tuned to a common protocol; quoting HGB is the conservative choice, since the logistic
vector read-out would make the gap 16.6 points and the text read-out 16.9. The deployed head appears as 0.7252 in the
P2 dump and 0.7257 in the frozen-backbone feature pass on the same 1{,}982 rows; the 0.05-point difference is a
code-path/bf16 difference, every cross-model contrast uses 0.7252, and the head comparison uses 0.7257 within its own
pass.

\begin{table}[!htbp]
\centering
\footnotesize
\caption{Teacher-analysis read-outs on the round-one test rows. These controls come from separate runs with different
readers and feature sets and are not recomputed under one protocol. The HGB on the teacher's recorded analysis vector is
the primary baseline for the 15.6-point content gap; the structured-plus-analysis HGB is the full-information-set
fair-baseline comparison; the cross rows train the same HGB on one analysis-vector source and evaluate it on the
other; the logistic text-versus-vector pair is the frozen-encoder control.}
\label{tab:teacherread}
\setlength{\tabcolsep}{3pt}
\begin{tabular}{llcc}
\toprule
Read-out & Features & Rows & Bacc \\
\midrule
HGB & teacher analysis vector (1024-d) & 1{,}998 & \textbf{0.9076} [0.894, 0.921] \\
HGB & teacher $\to$ generated vectors & 1{,}998 & 0.7421 [0.712, 0.768] \\
HGB & generated $\to$ teacher vectors & 1{,}998 & 0.7380 [0.710, 0.763] \\
HGB & structured + teacher analysis & 1{,}998 & 0.9044 \\
Logistic & teacher analysis text & 1{,}998 & 0.9205 \\
Logistic & teacher analysis vector (1024-d) & 1{,}998 & 0.9180 \\
\bottomrule
\end{tabular}
\end{table}

\paragraph{Round-one cross-reader and upstream checks.}
The two cross rows of Table~\ref{tab:teacherread} train the same HGB family on one analysis-vector source and
evaluate it on the other on the same 1{,}998 rows and labels: teacher-trained 0.7421, generated-trained 0.7380, against
the within-distribution 0.9076 and 0.7516. The upstream substitution in Section~\ref{sec:interface} injects the recorded
indicator query and summary at stage two with the checkpoint's teacher-forced-stage flag, holding the masks, state, news
and semantic history fixed; it is a forward pass only (no training, no teacher call). Scripts and JSON outputs are in
\texttt{v5\_analysis}.

\paragraph{Checkpoint roles.}
The deployed checkpoint is \texttt{multimodal\_5r\_hero\_s0\_best.pt} (training seed 0), used for all full-test
and closed-loop results and for the round-one sample fidelity. Four further training seeds of the same recipe (seeds
1--4) are used only for the seed-sensitivity and seed-aware numbers in Section~\ref{sec:interface} and this appendix.
The interface checkpoint (\texttt{d384base4}) is a two-layer variant of the same width (11.4M parameters); its seed-0
round-one diagnostics are 0.9242/0.7533 on the full 1{,}998 rows and 0.7526 on the 1{,}982 matched rows, and across
three seeds its matched self-analysis values are 0.7526/0.7589/0.7059 (mean 0.7391) against the retrain's 0.7522, so it
is not a reliable upper reference. In the 102-day closed loop it gives deterministic kurtosis 1.90 and volatility
89 bp (sampled: $-0.06$ and 119 bp), so it does not match the teacher's macro profile and is not deployed.

\paragraph{Stability.}
The deployed checkpoint is deterministic; its round-one numbers do not require seed averaging. The
closed-loop macro statistics use three simulator seeds. The HGB fair baseline is stable across its three seeds
(sd about 0.1 points), and the analysis-only reader family's three-seed sd is about 0.4 points. Most student
checkpoints are a single training seed (the behaviour-cloning control, the tied variant, the width variants). Four
recipes bound the training-seed scale. The analysis-only variant has three seeds: 0.8218/0.8279/0.8221 one-step
(sd 0.34) and 0.7419/0.7216/0.7412 autoregressive (sd 1.15), NLL 0.3700/0.3681/0.3700. The deployed recipe has five:
0.8347/0.8277/0.8370/0.8243/0.8317 one-step (mean 0.8311, sd 0.51), 0.7783/0.7772/0.7789/0.7712/0.7825
autoregressive (mean 0.7776, sd 0.41), NLL 0.3641/0.3562/0.3585/0.3605/0.3519, and round-one
0.7252/0.7348/0.7236/0.7342/0.7519 (mean 0.7339, sd 1.13). The no-semantic retrain has five seeds:
0.7842/0.7805/0.7919/0.7896/0.7778 one-step (sd 0.60), 0.7498/0.7561/0.7659/0.7535/0.7508 autoregressive (sd 0.65),
NLL 0.4157/0.4178/0.4227/0.4241/0.4203, and round-one 0.7485/0.7542/0.7536/0.7532/0.7517 (sd 0.23). The interface
checkpoint has three: seed-0 full-test P1 0.8192 and round-one 0.7526/0.7589/0.7059 (sd 2.9). The 2$\times$2 cells are
one training seed each, so their paired intervals include row noise but not training-seed noise. Row-paired intervals
condition on the trained runs; a two-level bootstrap over training seeds and date$\times$asset clusters resamples both.
As a descriptive yardstick we take twice the largest between-seed sd observed on any full-test variant of the same
decoder: 1.2 points one-step (retrain 0.60) and 2.3 autoregressive (analysis-only 1.15). With it, both schedule
removals clear the autoregressive value, the no-schedule removal clears the one-step value, and the analysis-only
one-step removal ($-1.3$) clears the 1.2 value by 0.1 point, less than the yardstick's own uncertainty, so we keep it
marginal; the 1.2-point control contrast sits at the one-step value, and the 0.82-point stage-loss effect, the
0.15-point tied margin and the interface--retrain differences are below it. Table~\ref{tab:seeds} collects the
replications.

\begin{table}[!htbp]
\centering
\footnotesize
\caption{Training-seed replication of the four recipes used in the comparisons. Balanced accuracy on the full test
window (one-step $n=9{,}956$; AR = within-day autoregressive, $n=9{,}910$) and on the 1{,}982 round-one self-analysis rows,
mean $\pm$ sd across the seeds. The interface checkpoint has no autoregressive run; its $n$ entry is one-step/round-one seeds (1/3), and its
round-one row is the matched-row value. The deployed checkpoint is seed 0 of the five deployed-recipe seeds.}
\label{tab:seeds}
\setlength{\tabcolsep}{1.5pt}
\resizebox{\columnwidth}{!}{%
\begin{tabular}{lcccc}
\toprule
Recipe & $n$ & One-step & AR & Round-one \\
\midrule
Deployed recipe & 5 & 0.8311 $\pm$ 0.0051 & 0.7776 $\pm$ 0.0041 & 0.7339 $\pm$ 0.0113 \\
No-semantic retrain & 5 & 0.7848 $\pm$ 0.0060 & 0.7552 $\pm$ 0.0065 & 0.7522 $\pm$ 0.0023 \\
Analysis-only schedule & 3 & 0.8239 $\pm$ 0.0034 & 0.7349 $\pm$ 0.0115 & -- \\
Interface checkpoint & 1/3 & 0.8192 (seed 0) & -- & 0.7391 $\pm$ 0.0291 \\
\bottomrule
\end{tabular}}
\end{table}

\paragraph{Not ablated.}
We do not separately ablate the stage tags, the visibility masks, or the encoder choice and dimension; the
design study varies the schedule, the stage losses and the head structure, and the no-semantic-history retrain removes
the semantic channel. The action head's size and offset are supervised ($0.01\cdot$MSE) and measured on the recorded
test calls: the size head's mean prediction is 0.83 of the teacher's mean size (Pearson 0.58, MAE 0.022) and its offset
magnitude is 1.06 of the recorded mean, so size is compressed. The deployed decoder takes the argmax side and the
\emph{median} size/offset; in closed loop this produces a thinner market (2{,}322 versus 13{,}484 lots per index per
day, with the student's volume decaying from about 6.1k lots per day in June to about 1.9k later), so the macro rows are
descriptive and not size-matched and a size-calibrated rerun is left to future work. On external baselines, the design
study includes the action-only behaviour-cloning control on the same architecture, data and split; a flat MLP on the
numeric context without any semantic channel reaches 0.7779 one-step (0.5000 for a constant predictor); and a flat-token
transformer with the same inputs and a matched parameter budget but no typed interface reaches 0.8177 one-step, 1.7
points below the staged student (Section~\ref{sec:interface}). We do not train a small language model or a CODI-style
continuous chain-of-thought baseline, and we have no second trace or second teacher realization; these are scope limits,
not withheld results. The flat comparison also lacks the stage losses and the schedule those stages enable, so we claim
the measured one-step benefit of the staged pipeline, not an additive decomposition of typing, stage supervision and
schedule.

\subsection{A twenty-seed capability sweep}
\label{sec:sweep}

The held-out check in Section~\ref{sec:deployment} uses one simulator seed. A student run is cheap enough that we can replace it with a
distribution: twenty seeds on each of the ten indices over the ten held-out days, i.e.\ 200 runs of one index each
(200{,}000 decisions; the unit is one index over the ten-day window, not a full ten-index run). The sweep takes 12.0
minutes wall-clock on three CPU processes (8.8 s per run; 29.2 minutes of process time) and zero LLM calls. The teacher side is the expensive part: one ten-day realization of the same ten
indices is about 9.1 minutes and 19{,}900 calls, so the same twenty realizations would require about 3.0 hours and
$4.0\times10^5$ calls before the analysis, roughly 15 times the student wall-clock for the same 20 realizations. The
student-side wall-clock excludes the one-off encoder pass over the trace, which is not part of the sweep. Table~\ref{tab:capability} reports the seed-level macro statistics:
kurtosis $-0.29 \pm 0.28$ (range $-0.74$ to $0.23$), volatility $146.9 \pm 7.5$ bp (134 to 163), and maximum daily
move $3.15 \pm 0.22$\% (2.89 to 3.71). The single-seed held-out values above ($-0.28$ and 137.6 bp) lie within 1.3 seed
standard deviations of these means; that is consistent with the single-seed values being representative, but it does
not test between-run teacher variance, which we do not estimate. The seed means are near the ten-index teacher path on the same window (kurtosis
$-0.48$, volatility 139.8 bp, maximum move 3.00\%): kurtosis differs by 0.19, volatility by 7.1 bp and maximum move by
0.15 pp, and the volatility sits 0.9 bp above the teacher's date-block upper bound (146 bp) while the maximum move sits
0.15 pp above the teacher path's observed maximum. A date-block bootstrap of a maximum cannot exceed the observed
maximum, so we do not treat the max-move interval as a pass/fail check and report the point difference only. We read
this as the same range rather than as equality. The seed spread is itself the useful output: differences of this
size across single runs are within simulator noise, and the surrogate makes the uncertainty statement practical.

\begin{table}[!htbp]
\centering
\footnotesize
\caption{Twenty-seed capability sweep on the ten-day held-out window (2023-11-02..2023-11-15): 200 deterministic
closed-loop runs (ten indices $\times$ twenty simulator seeds). Each seed contributes the mean over the ten indices;
columns are mean, sd, min, and max across the twenty seed-level means. The ten-index teacher recorded path on the same
window is $-0.48$ excess kurtosis, 139.8 bp volatility, and 3.00\% maximum move (date-block bootstrap intervals
$[-1.15,0.27]$, $[101,146]$ bp; the max-move interval $[2.15,3.00]\%$ is bounded above by the observed maximum by
construction and is not used as a pass/fail check); this is a single recorded teacher path on the same window.}
\label{tab:capability}
\setlength{\tabcolsep}{3pt}
\begin{tabular}{lcccc}
\toprule
Statistic & Mean & SD & Min & Max \\
\midrule
Excess kurtosis & $-0.29$ & 0.28 & $-0.74$ & 0.23 \\
Volatility (bp) & 146.9 & 7.5 & 134.1 & 162.9 \\
Max daily move (\%) & 3.15 & 0.22 & 2.89 & 3.71 \\
\bottomrule
\end{tabular}
\end{table}

\subsection{Compute--fidelity frontier}
\begin{figure}[!htbp]
\centering
\includegraphics[width=\columnwidth]{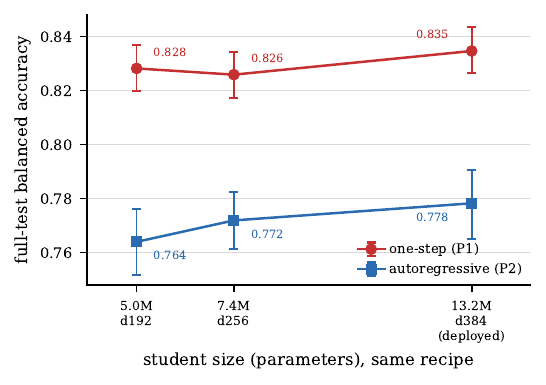}
\caption{Compute--fidelity frontier. The same recipe is trained at student widths 192, 256, and 384 (5.0M, 7.4M, 13.2M
parameters); points are full-test balanced accuracy for the one-step teacher-forced replay and the within-day
autoregressive rollout, with 95\% date$\times$asset cluster-bootstrap intervals. The autoregressive point estimates are
monotone in width, but the intervals overlap; the one-step point estimates are not monotone and also overlap.}
\label{fig:pareto}
\end{figure}

The same recipe at widths 192, 256, and 384 (5.0M, 7.4M, 13.2M parameters) gives one-step balanced accuracy 0.8282,
0.8259, 0.8347 and autoregressive 0.7640, 0.7719, 0.7783; the intervals overlap, so the ordering is descriptive. At
38\% and 56\% of the deployed parameter count the 5.0M and 7.4M students are 0.7/1.4 and 0.9/0.6 points below the
deployed model, so width is a usable compute knob rather than a one-dimensional scaling law.
Figure~\ref{fig:pareto} plots the frontier.

\subsection{Additional semantic-intervention probes}
\label{sec:probes}
\paragraph{Reliance on the semantic slots.}
On the full-test teacher-forced rows ($n=9{,}956$) we alter only the semantic-history input of the deployed
checkpoint. The unchanged pass gives 0.8344 (0.8347 in the main evaluation; the 0.03-point difference is a different
batch pass). Masking the semantic slots lowers full-test balanced accuracy to 0.7738 ($-6.1$ points) and round one from
0.9098 to 0.7218 ($-18.8$; 24.5\% of decisions flip); at round one the only active slot is the current analysis, so this
is a current-analysis effect. Replacing the active slots with random 1024-d vectors costs more, 0.7402 ($-9.4$;
round one 0.6761). Setting the current-analysis half of each active slot to its normalised mean leaves later rounds
almost unchanged (3.4--6.2\% flips) but collapses round one to 0.6149 ($-29.5$; 31.6\% flips), whereas setting the
rationale half to its mean costs 3.4 points at round one and 2.3 overall. These are test-time interventions on a student
trained with the channel, so they measure reliance, not what a student can do without it; the latter is the
no-semantic retrain of Table~\ref{tab:design} ($+5.0$ teacher-forced, $+2.3$ autoregressive over the five-retrain mean)
and the content controls of Appendix~\ref{sec:shuf}. The two readings are consistent: the trained student relies on an
input that, when it must generate it itself at round one, it does not convert into an accuracy gain.

\paragraph{The analysis vector is a live input.}
We intervene directly on the semantic input. On the 1{,}995 round-one test decisions for which same-side and
opposite-side donors both exist, we replace the recipient's current analysis vector with another agent's analysis
vector from the same asset-day and re-run the student; the structured inputs, persona, and the rest of the context are
unchanged. An opposite-side donor (a same-day agent whose recorded action is opposite to the recipient's) flips the
action on 64.7\% of decisions [59.8, 69.3] and moves the buy probability by 0.61 on average in absolute value; the share
of decisions on the donor's side rises from 8.7\% before the swap to 72.2\% after it ($+63.5$ points [58.4, 68.2]). A
same-side donor's analysis flips 12.8\% [10.9, 14.8] and moves the probability by 0.15, and a random same-day donor
gives 31.3\% [28.3, 34.3]. The same-side/opposite-side contrast holds the act of substitution fixed and varies only the
donor's side, so the analysis vector is not a generic input the action ignores. The donors are observed, not randomised,
and a donor vector also encodes the donor's persona, account state and private information; we therefore read this as
input sensitivity of the analysis channel rather than as a causal effect of semantic content alone. The unchanged pass
on these rows has balanced accuracy 0.9096, matching the 0.9098 reported above on the full 1{,}998 contexts.

\subsection{Student architecture and training}
Each free-form teacher output is encoded once, offline, with a frozen Qwen3-Embedding-0.6B encoder into a 1024-d unit
vector. The deployed student is a Transformer encoder with three layers, width 384, eight attention heads,
feed-forward width $4d$, dropout 0.1, a causal mask, and per-modality input heads for persona, retrieved news, state,
extra context, previous-round parameters, previous actions, round history, day history, indicator values, competition,
and day-cumulative features. Six stage heads produce the news query, the indicator query, the indicator summary, the
day analysis, the round rationale, and the action. Stage heads are masked by attention prefixes: the news query attends
to the persona, the analysis head to persona, news, indicator summary, account state and day history (never to same-day
rounds), and the rationale and action heads to the full context. The loss is a cosine loss on the four semantic stages,
$0.5\cdot$BCE $+\,0.05\cdot$MSE on the indicator query, cross-entropy on the action side, and $0.01\cdot$MSE on order
size and offset. Each stage has its own linear output head back into the shared 1024-d semantic space. Training uses
Adam with learning rate $3\times10^{-4}$, batch 256, bf16, at most 30 epochs with patience 7 (best epoch 13), and takes
1{,}658 s on one 8-GB consumer GPU (RTX 4060 Laptop). Scheduled self-history replaces the analysis slot with the
student's own vector with probability 0.5 and the rationale/action-history slots with probability 0.35. The encoder is
applied once, offline, and its cost is part of trace preprocessing rather than of the student training.

\subsection{Deployment economics}
\begin{table}[!htbp]
\centering
\footnotesize
\caption{Deployment economics. The first run of the surrogate is not cheaper than a teacher run: it consumes the
recorded trace (203{,}100 calls, 691.5M prompt tokens, about 93 min), one frozen-encoder pass over the trace text,
1{,}658 s of student training and about 2 min of simulation. The amortisation is on re-runs: each further 102-day
realization costs about 2 min and no LLM calls. Student counts exclude the frozen 0.6B embedding encoder, which is
applied once, offline, and is not in the decision loop. The 119{,}880 replay calls are evaluation-only and are not in
any column. New text outside the pre-encoded corpus would need one encoder pass per stage output, so the zero-call row
is an inference-loop statement.}
\label{tab:deploy}
\setlength{\tabcolsep}{3pt}
\begin{tabular}{lcc}
\toprule
 & Teacher run & Student \\
\midrule
LLM calls & 203{,}100 & 0 after the trace \\
Prompt tokens & 691.5M & 0 \\
Trace and encoder & produced by the run & one-time input, 1 pass \\
Training & -- & 1{,}658 s, 8-GB GPU \\
First 102-day run & 93 min & trace + 1{,}658 s + 2 min \\
Re-run, 102 days & 93 min & $\approx$2 min \\
Parameters / size & -- & 13.2M / 52.7 MB \\
Latency, per decision & sec--min & 0.14--10.7 ms \\
\bottomrule
\end{tabular}
\end{table}

\subsection{Compute}
The recorded teacher run required 203{,}100 LLM calls, 691.5M prompt tokens, and about 93 minutes. A 102-day, ten-index deterministic student run takes about 115 s wall-clock (346 s for the three seeds in Table~\ref{tab:macro}), so an end-to-end first run costs the trace plus 1{,}658 s of student training plus about 2 min of simulation, while a re-run costs about 2 min; one encoder pass per stage output is needed only for text outside the pre-encoded trace. The student has
13{,}175{,}439 parameters (52.7 MB), trains in 1{,}658 seconds, and runs at 0.14 ms per decision at batch 256
(0.59 ms at batch 20, 10.7 ms per decision); the deployed loop performs one analysis pass per agent-day and one pass per
round. The re-sampling analysis adds 119{,}880 teacher replay attempts (successful-draw means 19.97/19.98, about 15 minutes
wall-clock at the replay harness's request concurrency) but no student-side training and is not part of the deployed
simulator; because these are one-stage decision and analysis completions, the wall-clock total is not comparable to the
203{,}100-call, 93-minute trace run's per-call rate. No language-model calls are made inside the
deployed student's inference; the teacher replay is evaluation-only. The anonymised supplementary archive contains the
training and evaluation code, the configuration files, the metric JSON behind every table, the derived row-level
predictions, and a script that regenerates Tables~\ref{tab:fidelity}--\ref{tab:deploy} from those JSON files; the
recorded trace, the frozen encoder weights and the teacher replay outputs are not redistributed because they are
subject to their original terms. The archive is organised as \texttt{code/}, \texttt{configs/}, \texttt{metrics/}
and \texttt{reproduce/}, and the table-regeneration script is \texttt{reproduce/make\_tables.py}.

\subsection{Data sources, teacher prompts, and encoder pinning}
\paragraph{Real-market reference.}
The real-market row of Table~\ref{tab:macro} is computed from the ten Chinese industry index series that the
simulator ships for the same dates (\texttt{data/real\_market.csv}; daily close levels for the ten Tushare industry
indices, 2023-06-15 to 2023-11-15). For each index we take daily simple returns and average over the ten indices the
excess kurtosis (moment estimator), volatility in basis points, maximum absolute daily return, and the ARCH(1) lag-1
autocorrelation of squared returns; the same definitions are used for every row of the table. The real series is an
external reference that neither the teacher nor the student is trained on.

\paragraph{Teacher prompts and stage tags.}
The teacher is \texttt{deepseek-chat} at temperature 0.9, top-p 0.9, frequency and presence penalties 0.5 and
\texttt{max\_tokens} 4096. Prompts are built by \texttt{trader/prompts.py} and \texttt{trader/trading\_agent.py} and
stored per call in the trace's \texttt{05\_prompt\_traces}: each decision prompt carries the round header, the account
and order-book state, the retrieved news and indicator summary, and the decision template, and each analysis request
carries the retrieved news and indicator evidence. The student's six stage tags follow the recorded message roles
(Table~\ref{tab:mask}); the visibility matrix is the student's implementation, not a template-level audit of the
teacher prompt, which contains formatting and retry instructions that we do not model. The stored prompts are the ones
used for all encoded targets; no prompt was regenerated for this paper.

\paragraph{Encoder.}
Every free-form stage output is encoded once, offline, with the frozen \texttt{Qwen/Qwen3-Embedding-0.6B} model
(1024-d, last-token pooling followed by L2 normalisation, bf16) loaded from
\texttt{llm2nn/data/models/Qwen3-Embedding-0.6B}. The local snapshot carries no revision hash, so we pin the model name
and the local path rather than a commit. The encoder is not trained and is not called in the deployment loop.

\paragraph{Replay denominator and secondary comparisons.}
The replay reference exists at round one only: we did not run later-round teacher replays, so no full-window or
per-round $R$ is reported, and the autoregressive comparison is relative rather than a fidelity statement. The
parameter-matched head refits, the stage-two substitution, the donor probes, the cross-reader cells and the
2$\times$2 design contrasts are exploratory and not multiplicity-corrected; the primary comparison is the five-seed
deployed-versus-retrain contrast.